\documentclass[sn-mathphys-num]{sn-jnl}% Math and Physical Sciences Numbered Reference Style 
\usepackage[utf8]{inputenc}

\usepackage{graphicx}%
\usepackage{multirow}%
\usepackage{amsmath,amssymb,amsfonts}%
\usepackage{amsthm}%
\usepackage{mathrsfs}%
\usepackage[title]{appendix}%
\usepackage{xcolor}%
\usepackage{textcomp}%
\usepackage{manyfoot}%
\usepackage{booktabs}%
\usepackage{algorithm}%
\usepackage{algorithmicx}%
\usepackage{algpseudocode}%
\usepackage{listings}%

\usepackage[numbers]{natbib}

\usepackage{float}
\restylefloat{table}
\usepackage{multicol}
\usepackage{stmaryrd}
\usepackage{verbatim}
\usepackage[T1]{fontenc}
\usepackage{arydshln}
\usepackage{comment}

\theoremstyle{thmstyleone}%
\theoremstyle{thmstyletwo}%

\theoremstyle{thmstylethree}%

\begin{document}

\title[Leveraging Physiological Signals to Predict Exam Outcomes with Machine Learning]{Leveraging Physiological Signals to Predict Exam Outcomes with Machine Learning}


\author[1]{\fnm{Lala} \sur{Yamazaki}}\email{ly2z@mtmail.mtsu.edu}

\author*[2]{\fnm{Ramchandra} \sur{Rimal}}\email{ramchandra.rimal@mtsu.edu}

\affil*[1, 2]{\orgdiv{Department of Mathematical Sciences}, \orgname{Middle Tennessee State University}, \orgaddress{\street{1301 E Main St}, \city{Murfreesboro}, \postcode{37132}, \state{TN}, \country{USA}}}

% \affil[2]{\orgdiv{Department}, \orgname{Organization}, \orgaddress{\street{Street}, \city{City}, \postcode{10587}, \state{State}, \country{Country}}}

% \affil[3]{\orgdiv{Department}, \orgname{Organization}, \orgaddress{\street{Street}, \city{City}, \postcode{610101}, \state{State}, \country{Country}}}


\abstract{

This study investigates the application of machine learning models to predict exam outcomes using physiological data collected during examination sessions. Physiological stress indicators, including electrodermal activity, heart rate, and skin temperature, were analyzed to uncover their association with academic performance. A variety of machine learning approaches were employed, ranging from standard models like logistic regression, random forest, and support vector machines to more advanced architectures, including transformers, long short-term memory (LSTM), and gated recurrent unit (GRU) models. This diversity aimed to capture the complex interactions within the data effectively. A key focus was assessing the adaptability of transformers in processing numerical data and evaluating their performance in this novel context. Standard performance metrics, such as accuracy, precision, recall, and F1-score, were used to compare model efficacy. The experimental results demonstrate that while deep learning models generally excel at capturing complex relationships in physiological data, simpler models like random forests can sometimes achieve superior performance while offering computational efficiency and interpretability. Furthermore, transformers demonstrated notable versatility, showcasing performances comparable to those of the LSTM and GRU models. This research underscores the importance of experimenting with a broad class of models that align with the objectives of the problem at hand, balancing precision, efficiency, and interpretability. By elucidating the relationships between physiological signals and academic performance, this study contributes to understanding stressors affecting students' mental health. It further promotes leveraging physiological data to enhance student well-being and academic outcomes. }




\maketitle
%{\large\bf Orange: highlighted for review;\\ Red: Feedback;\\ Blue- Suggestion/Revised version }
\section{Introduction}\label{sec1}

%\The Introduction section, of referenced text \cite{bib1} expands on the background of the work (some overlap with the Abstract is acceptable). The introduction should not include subheadings.%\

The rapid advancements in machine learning (ML) have introduced powerful tools for predictive analytics and complex data interpretation across various fields. Among these innovations, the long short term memory(LSTM), gated recurrent unit(GRU), and, most recently, transformer models have achieved groundbreaking success, particularly in natural language processing, due to their ability to capture intricate patterns and dependencies within data. In addition, transformers have achieved impressive results in sequential tasks like time series forecasting, demonstrating similar or sometimes better performance compared to LSTM and GRU models~\cite{bib21, bib42}. However, their adoption in general machine learning tasks or other complex numerical data remains limited~\cite{bib22}. Furthermore, the parallelizability of the transformer models makes it more compelling than LSTM and GRU's. This gap presents an exciting opportunity to investigate the potential of transformers in a novel context: analyzing physiological responses to stress and predicting academic outcomes based on numerical physiological data. In this study, transformers are not only central but are also specifically modified and adapted to handle classification tasks with numerical data, offering a unique contribution to both machine learning and applied physiological research.
%%%%%%%%%%%%%%%%%%%%%%%%%%%

The main motivation for this research is to accurately predict the exam outcomes based on physiological signals, such as electrodermal activity, heart rate, and skin temperature, to name a few, collected during exam sessions. We further intend to explore how these indicators of stress measured through signals are associated with academic performance. Traditional machine learning models, including logistic regression, support vector machine (SVM), and random forest, have commonly been used in physiological data analysis, offering interpretability and practical insights~\cite{bib45, bib46, bib47}. However, these models struggle to capture the complex relationships and variability of some real-world datasets, limiting their predictive power in tasks that require nuanced pattern recognition and sequential data analysis~\cite{bib48, bib49}. By contrast, deep learning architectures, particularly transformers and recurrent neural networks such as LSTM and GRU models, are designed to handle sequential, complex data, potentially providing higher predictive accuracy. On the flip side, these models offer limited statistical inference, so they are not well suited when the goal of the study is to understand predictors and response relationships. This study aims to evaluate the effectiveness of both classes of models, with a particular focus on transformers, in predicting academic outcomes using data collected in a high-stress real-world environment.
%%%%%%%%%%%%%%%%%%%%%%%%%%%

Mental health has become a significant concern, especially for working-class families~\cite{bib50, bib51}. By understanding how stress impacts academic performance, we can develop tailored support systems and interventions for students. This approach aims to enhance cognitive performance and mental health through adaptive, data-driven methods. This research has the potential to contribute to mental health and well-being in educational settings and any context where mentally demanding work occurs. Additionally, by adapting transformer models to analyze physiological data and evaluating the effectiveness of various machine learning models for classification tasks, this study advances machine learning while providing a foundation for practical tools and interventions in high-stress work environments.
%%%%%%%%%%%%%%%%%%%%%%%%%%%

Finally, the previous research published with this data utilizes only skin electrical conductance measured by electrodermal activity(EDA), one of the eight physiological features collected. %The study divides the data signal into three parts: start, middle, and end. It then defines the difference signal and computes the mean and variances of these parts and the difference signal. It also calculates the ratio of the average of mid to the sum of averages of start and end. 
Nine engineered features by partitioning, differencing, and calculating ratios from EDA were utilized as input features and predicted grades for the exams, with the highest accuracy of 80\%  for the final exam~\cite{bib14}. Our work utilizes all the useful collected features, performs a comprehensive analysis to improve predictive performance, and discusses each feature's contribution to performance. More specifically, using a multivariate dataset that captures various physiological responses during exam sessions, this study compares transformers with traditional and deep learning models, including logistic regression, SVM, random forest, LSTM, and GRU. This study also explores whether transformers, modified explicitly for classifying numerical data, can deliver enhanced predictive accuracy, given their recognition as state-of-the-art models across various applications.

This paper is organized as follows. Section~\ref{sec2} begins with a discussion of related work, providing context for the study by reviewing existing literature on machine learning and deep learning models for physiological data analysis, highlighting the gaps this research aims to address. Section~\ref{sec3} introduces the dataset, detailing its features, the data collection process, and their relevance in examining stress responses during cognitive challenges. This section also outlines the preprocessing steps to prepare the data for analysis. Section~\ref{sec4} outlines the predictive methodology, providing a brief overview of the logistic regression, SVM, random forest, LSTM, and GRU models and a detailed examination of transformer models and their customization for classification tasks involving numerical data. Section~\ref{sec5} details the experimental design, model construction, and hyperparameter tuning and results, providing a comprehensive comparative analysis of each model's predictive performance. Section~\ref{sec6} discusses the ethical considerations and practical implications of the study, ensuring transparency,  and accountability. Section~\ref{sec7} wraps up the paper by highlighting the main findings, addressing their implications for future research, and examining potential applications. The detailed  results of the statistical analysis are outlined in the appendix Section~\ref{secA1}.\\
%%%%%%%%%%%%%%%%%%%%%%%%%%%

\section{Related Work}\label{sec2}

The continuous evolution of ML and in particular, deep learning (DL) techniques, fueled by improvements in computational power and the increasing accessibility of diverse datasets, has enabled the development of sophisticated models for forecasting cognitive states and performance outcomes. Recent studies have increasingly explored physiological signals to understand and predict cognitive states under stress, particularly through ML and DL models applied to real-world scenarios. This section reviews developments in applying these models to analyze physiological data for cognitive performance prediction and stress assessment, highlighting their strengths, limitations, and the potential of novel approaches.

Numerous ML models, including SVM, random forests, and gradient boosting, have shown success in classifying stress from physiological signals, often outperforming traditional statistical methods in this context~\cite{bib15, bib16, lazarou2024predicting}. These methods exploit temporal correlations in physiological data, such as heart rate variability (HRV) and EDA, critical stress, and cognitive load indicators. Wijsman et al.(2011) develop a wearable sensor system to measure physiological signals and detect mental stress. Out of nineteen features derived from ECG, respiration, skin conductance, and EMG signals, nine were selected based on correlation analysis. Principal component analysis reduced these features to seven principal components, enabling classifiers to achieve approximately 80\% accuracy in distinguishing stress from non-stress conditions~\cite{wijsman2011towards}.  Studies by Panicker and Gayathri (2019) and Martinez-Ríos et al. (2021) demonstrate the utility of ML for handling such data, particularly when enriched with additional socio-demographic factors~\cite{bib15, bib16}.  However, the effectiveness of these models varies significantly based on feature selection and data preprocessing, indicating a need for a unified framework that maximizes predictive power across diverse input features.

Deep learning models, especially those geared towards sequence modeling like LSTM networks and convolutional neural networks, have further improved prediction accuracy in tasks requiring memory retention over time. Bota et al. (2019) and Rush et al. (2018) applied these models to physiological signals in healthcare settings, noting the benefits of DL in managing complex, continuous data streams~\cite{bib17, bib18}. \cite{agarwal2023physiological} performed regression analysis using the same data as ours but  only considering three stressors, such as skin temperature, electrodermal activity, and heart rate, in predicting the students' grades over three examinations. They utilize common regression models, including  more complex deep neural networks, and convolutional neural networks. They augmented the data and preprocessed data differently by calculating mean features across three-time intervals and using them as the features. Their experimental results also demonstrate random forest outperforms other models experimented in root mean squared error. Despite their success, these works  often lack interpretability and require substantial training data, limiting their practical application in small-scale, real-world datasets like those used in educational or high-stress environments.

Additionally, hybrid models combining ML and DL approaches have demonstrated enhanced robustness in tasks requiring multidimensional feature integration. Studies by Petrescu et al. (2021) applied ensemble approaches to classify emotions based on physiological data, achieving high accuracy in classifying fear—a high-arousal state relevant for cognitive performance under exam stress~\cite{bib19}. However, these hybrid models frequently emphasize single physiological indicators rather than integrated multivariate data, which could provide a richer understanding of cognitive states under stress. In their editorial, Tripathy, Paternina, and de la O Serna (2022) highlighted the application of both ML and DL in physiological signal analysis, particularly emphasizing the potential of DL techniques like transformers to capture complex temporal dependencies across multimodal data streams~\cite{bib20}. Their review underscores the advantage of transformers for tasks requiring continuous and long-term physiological monitoring.

A benchmark research for our work is based on the Wearable Exam Stress Dataset by Amin, Wickramasuriya, and Faghih (2022), which captures a comprehensive array of physiological signals, including EDA collected during real-world examinations~\cite{bib14}. Previous analyses of this dataset have predominantly focused on individual features, primarily electrodermal activity, to predict exam grades. These studies utilized solely skin electrical conductance measured through EDA, one of the eight physiological features collected, and engineered features based on this singular measurement. The dataset is divided into three segments: start, middle, and end. Researchers define the difference signal and compute the means and variances of these segments alongside the difference signal. Additionally, they calculate the ratio of the average of the middle segment to the sum of the averages of the start and end segments. Nine engineered features derived from EDA were then used as input features to predict exam grades, achieving a maximum accuracy of 80\% for the final exam~\cite{bib14}. However, incorporating the full range of physiological signals  can produce more insightful and accurate predictive models of cognitive performance under stress, addressing a gap that our study aims to fill by leveraging these multimodal recordings.

While prior research has contributed significantly to our understanding of physiological signals in predicting cognitive performance, this paper aims to provide a more comprehensive and in-depth analysis by utilizing advanced ML and DL approaches. By leveraging a state-of-the-art transformer model, recognized for its capability to capture intricate dependencies within complex datasets, in conjunction with other promising ML and DL techniques, this study aspires to enhance both predictive accuracy and interpretability. Adopting a multivariate analysis framework, this research seeks to uncover nuanced insights into cognitive performance under real-world stress conditions, thus deepening our understanding of this complex interplay. Specifically, our study pursues the following initiatives to address the aforementioned challenges and bridge existing research gaps: (a) identifying and selecting all pertinent features from the collected data, (b) adapting the transformer architecture for the classification modeling of general data, (c) developing an integrated and customizable computational framework utilizing various machine learning techniques, (d) experimenting with multiple model configurations with varying levels of complexity, (e) designing a robust data-driven approach for hyperparameter tuning and model selection, and (f) obtaining superior results with  96\% accuracy over 80\% previously. The findings enhance predictive modeling capabilities while offering valuable insights to inform future research and drive advancements in educational applications and mental health.
%%%%%%%%%%%%%%%%%%%%%%%%%%%

\section{Data}\label{sec3}

The dataset utilized in this research is a carefully assembled collection designed
to capture the physiological responses of students under real-world cognitive stress
conditions, specifically during exam situations. This dataset, named ”A Wearable
Exam Stress Dataset for Predicting Cognitive Performance in Real-World Settings,”
addresses a notable gap in the current body of knowledge, which has largely been
built on controlled, laboratory-based studies. By focusing on real-world scenarios,
this dataset offers a unique opportunity to explore how stress manifests in everyday
academic environments and its subsequent impact on student performance~\cite{bib23}.
%%%%%%%%%%%%%%%%%%%%%%%%%%%

\subsection{Data Collection}\label{subsec1}
The data for this study was obtained from the open-access repository on PhysioNet and was collected from 10 college students using the Empatica E4 wristband~\cite{bib23}. This FDA-approved wearable device is known for its precision in capturing physiological signals. Participants wore the E4 wristband during three key exam sessions: Midterm 1, Midterm 2, and the Final exam. These sessions were conducted in typical academic settings, allowing for natural variations in stress levels across standard evaluation periods~\cite{bib23}. Each midterm exam lasted 1.5 hours, providing an opportunity to capture physiological data under common exam conditions. The final exam, with a longer duration of 3 hours, allowed for the observation of stress responses over an extended period, providing valuable data on sustained cognitive and emotional pressure. The E4 wristband recorded five key physiological parameters relevant to stress analysis, briefly explained below.\\
%%%%%%%%%%%%%%%%%%%%%%%%%%%
\noindent
\textbf{Electrodermal Activity (EDA)} measures the skin's electrical conductance, which varies with sweat gland activity influenced by psychological or physiological arousal. It is commonly used to assess emotional responses and stress levels as it reflects sympathetic nervous system activity.\\
 \textbf{Heart Rate (HR)}  refers to the number of heartbeats per minute and provides insights into the body's cardiovascular response to stress or relaxation. Changes in HR are often associated with physiological and emotional responses to external stimuli. \\
 \textbf{Blood Volume Pulse (BVP)} captures the amount of blood flow in peripheral blood vessels, which can vary with heart rate and vascular changes. This data helps assess cardiovascular responses such as pulse rate and the effects of stress or relaxation on circulation.\\
  \textbf{Skin Surface Temperature (TEMP)}  indicates the body's external temperature, which can change due to factors like stress or environment. A decrease in skin temperature can sometimes correlate with heightened stress or arousal. \\
  \textbf{Accelerometer Data (ACC)} captures body movement by measuring acceleration along three axes, often used to detect physical activity and posture. ACC helps differentiate physiological changes due to movement from emotional or mental stress responses.
%%%%%%%%%%%%%%%%%%%%%%%%%%%

\subsection{Data Preprocessing}\label{subsec2}

The preprocessing pipeline begins with setting up the environment. Raw physiological data for EDA, HR, BVP, TEMP, and ACC are read and standardized to ensure consistency across files. Metadata fields are then added to enrich the dataset with contextual information. This includes each student's unique identifier, exam session details, and measurement type. This metadata aids in data organization and facilitates targeted analysis across different individuals and exam conditions.

\subsubsection{Data Pivoting and Transformation}

To prepare the data for sequential modeling, the raw data is transformed from a long format, where each row corresponds to a single measurement, to a wide format, with each row representing the measurements for a specific timestamp. Missing values, common in real-world physiological datasets due to sensor limitations or incomplete data collection, are identified. To maintain dataset integrity, rows with missing values are removed.

\subsubsection{Grade Integration and Normalization}

Student performance data, specifically grades from each exam session, is integrated by matching each student's physiological data with their corresponding grades using student IDs and exam sessions. Grades are normalized to a percentage scale, with final exam scores scaled from a maximum of 200 points to a 100-point scale. These normalized grades are then assigned values from 0 to 4, which are subsequently categorized into letter grades ranging from A to F. The Grade column, containing raw scores, is later dropped from the final dataset to standardize the analysis and focus on the categorized grades.

In addition to physiological features, the final dataset includes binary indicators for exam participation: Midterm 1 Taken, Midterm 2 Taken, and Final Taken. These columns indicate whether a student participated in the corresponding exam session, with a value of 1 representing participation and 0 indicating non-participation. The final dataset, therefore, retains only the participation indicators, physiological features, and the categorized grade for training and evaluation.

\subsubsection{Class Imbalance Handling and Final Dataset}

To address class imbalance in the grade categories, an undersampling technique is applied. This process reduces the number of samples in the majority classes to match the minority class, resulting in a balanced dataset for model training. Each grade category was balanced with 65,731 observations.

The final processed dataset contains 328,655 samples, structured with predictor and target variables suitable for machine learning analysis. Table~\ref{Table: features_roles} provides detailed descriptions of the predictor and target variables.

%%%%%%%%%%%%%%%%%%%
\begin{table}[H]
\centering  
\caption{Features and their roles in the dataset}\label{Table: features_roles}
\begin{tabular}{c|c|c}
\textbf{Feature} & \textbf{Description} & \textbf{Role} \\ 
\hline
ACC & Accelerometer Data & Predictor \\ 
BVP & Blood Volume Pulse & Predictor \\ 
EDA & Electrodermal Activity & Predictor \\ 
HR & Heart Rate & Predictor \\ 
TEMP & Skin Temperature & Predictor \\ 
Midterm 1 Taken & Participation in Midterm 1 & Predictor \\ 
Midterm 2 Taken & Participation in Midterm 2 & Predictor \\ 
Final Taken& Participation in Final Exam & Predictor \\ 
Grade Category & Normalized grades & Target \\ 
\end{tabular}
\end{table}
%%%%%%%%%%%%%%%%%%%

Predictor Variables include physiological features such as ACC, BVP, EDA, HR and TEMP. Exam participation indicators (Midterm 1 Taken, Midterm 2 Taken, and Final Taken) are also included to capture contextual exam-related influences. The target variable is the normalized grade category. Midterm exams were graded out of 100, while the final exam was graded out of 200. Final exam grades were normalized to a percentage scale by dividing scores by 200 and multiplying by 100. The categorization of grades, along with their corresponding grade categories and the number of observations per category, is detailed in Table~\ref{Table:grade_calculation}.

\begin{table}[H]
\centering  
\caption{Grade calculation, normalization, and categorization}\label{Table:grade_calculation}
\begin{tabular}{c|c|c|c}
\textbf{Grade Category} & \textbf{Grade} & \textbf{Percentage Range} & \textbf{No. of Observations} \\ 
\hline
0 & A & 90\% and above & 65,731 \\ 
1 & B & 80\% to 89\% & 65,731 \\ 
2 & C & 70\% to 79\% & 65,731 \\ 
3 & D & 60\% to 69\% & 65,731 \\ 
4 & F & Below 60\% & 65,731 \\ 
\hline
\multicolumn{4}{l}{\textbf{Normalization Formula:} $\text{Normalized Score} = \frac{\text{Exam Final Score}}{200} \times 100$}
\end{tabular}
\end{table}

The final dataset aligns five physiological measurements by timestamps and includes exam-related indicators and normalized grade categories.

\subsection{Exploratory Data Analysis}\label{subsec3}

As part of the exploratory data analysis, the distributions of physiological measurements were examined to uncover patterns indicative of stress and focus during exam sessions. Histograms and kernel density estimates were used to visualize the data, providing insights into the variation and density of each physiological signal across students.

\begin{figure}[H] % force exact placement
    \centering
    \includegraphics[width= 1.2\textwidth]{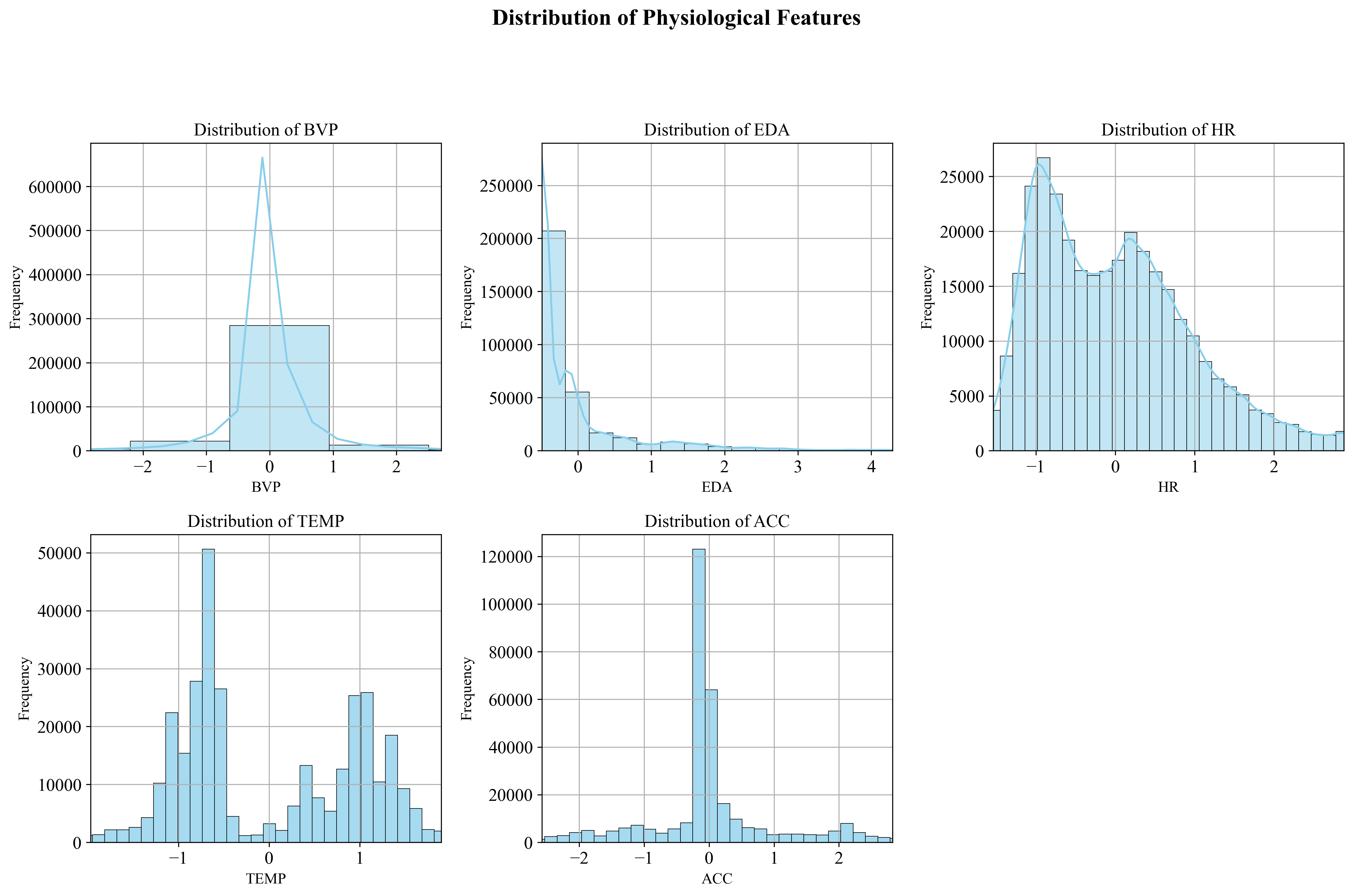}
    \caption{Distribution of physiological features: BVP (top left), EDA (top middle), HR (top right), TEMP (bottom left), and ACC (bottom right) }
    \label{fig:distr_feat}
\end{figure}

Based on Figure~\ref{fig:distr_feat}, the ACC data exhibits a sharp peak near zero, indicating minimal movement, suggesting periods of stillness likely associated with concentration. The smaller peaks and broader spread reflect intermittent activity, such as brief adjustments or fidgeting, which may be linked to shifts in focus or stress levels.

The BVP distribution features a pronounced central peak, indicating baseline cardiovascular activity, while the spread around this peak suggests variability in blood volume pulse, likely tied to moments of arousal or stress during the exam. Similarly, the EDA data demonstrates a steep initial peak at baseline arousal levels, with a tapering spread and occasional spikes that capture episodes of heightened stress or physiological activation.

The HR distribution shows multiple peaks and a wide spread, reflecting both commonly observed heart rate levels and elevated rates, likely associated with periods of stress or cognitive effort. Lastly, the TEMP data reveals multiple distinct peaks, indicating variations in skin temperature that may correspond to stress-induced thermoregulation.

\section{Methodology}
\label{sec4}

This study develops predictive models to forecast students' academic performance based on physiological signals collected during exam sessions. These signals are believed to reflect stress and cognitive load, which may influence grades. By employing a range of machine learning models, the study aims to enhance prediction accuracy and explore underlying patterns that link physiological responses to performance. The primary focus is leveraging classification models to analyze these signals and generate insights into how stress-related indicators relate to academic achievement.

\subsection{Logistic Regression}
\label{subsec5}

Logistic regression is a statistical model commonly used for classification tasks where the target variable is categorical. It models the relationship between one or more predictor variables and a binary dependent variable by predicting the probability of a particular outcome~\cite{bib32}. At the core of logistic regression is the logistic (sigmoid), function, which maps predicted values to probabilities between 0 and 1~\cite{bib33}. This probability is then used to classify an input into one of two categories based on a threshold.

This study uses multinomial logistic regression—an extension of logistic regression to a multi-class setting—to predict students' academic performance across several grade categories. This approach is appropriate to the chosen multivariate dataset, allowing the estimation of probabilities for each grade category based on physiological and exam-related features~\cite{bib35,bib36}. Multinomial logistic regression calculates distinct probabilities for each grade category by estimating separate coefficients for each class relative to a baseline. To handle the multi-class structure, the softmax function normalizes the probabilities, ensuring they sum to 1. Logistic regression's classification benefit lies in its simplicity and interpretability, making it particularly advantageous for identifying patterns and relationships in datasets with limited complexity. One notable advantage of logistic regression is its ability to provide feature importance, allowing insights into the relative contribution of each predictor to the classification outcomes. This advantage aligns well with the multinomial logistic regression framework, which provides a structured and interpretable method for classifying students into grade categories~\cite{bib34}.

\subsection{SVM}
\label{subsec5}

SVM, introduced by Cortes and Vapnik in 1995, is a supervised learning method for classification and regression tasks~\cite{bib29}. SVMs work by identifying a hyperplane that maximally separates data points of different classes, usually in a high-dimensional space, effectively creating a boundary with the largest possible margin from the nearest data points of each class. Maximizing this margin is central to SVMs, as it aims to increase the model's generalization capacity, allowing it to perform well on unseen data~\cite{bib29}.

In SVM, kernel functions such as linear, polynomial, and radial basis function (RBF) kernels are used to transform data that may not be linearly separable in its original space. This transformation enables SVM to find complex decision boundaries by mapping input features into higher dimensions, where a linear hyperplane can better separate classes. SVMs have shown effectiveness for numerical data classification due to their robustness and ability to handle high-dimensional data. For instance, in fields like finance and biology, SVMs have been used for numerical data classification tasks, such as predicting stock trends or classifying patient health status based on physiological measurements, leveraging the capacity of SVMs to learn and generalize from numerical features effectively~\cite{bib30,bib31, martinez2021review}.

\subsection{Random Forest}
\label{subsec6}

Random forest, introduced by Breiman in 2001, is an ensemble learning method that constructs multiple decision trees during training and combines their outputs to improve prediction accuracy. This method excels in classification and regression tasks by aggregating individual tree predictions. For classification problems, random forest employs majority voting across trees to determine the output class. By training each tree on random subsets of the dataset and selecting a subset of features randomly at each split, each tree in the forest learns unique patterns, leading to random forest reducing overfitting and enhancing generalization. This approach improves the model's robustness and efficiently addresses the bias-variance tradeoff~\cite{bib37}.

The model's ability to handle numerical data where relationships between variables are often complex and non-linear is a significant advantage. Like logistic regression, random forest provides feature importance, helping to identify the most influential variables in a dataset. This capability allows researchers to gain meaningful insights into predictor contributions and prioritize critical features. The model's resilience to noisy and variable data and its adaptability to imbalanced datasets further underscores its suitability for tasks requiring robust classification. For example, it has shown effectiveness in predicting academic success across exam sessions, where individual physiological differences pose significant variability~\cite{bib38, martinez2021review, ahmad2023framework}.

\subsection{LSTM}
\label{subsec7}

LSTM model is an improved recurrent neural network (RNN) designed to handle sequential data and mitigate the issue of vanishing gradients—a common challenge in traditional RNNs. Proposed by Hochreiter and Schmidhuber in 1997, LSTM networks are structured with memory cells that include three gates: the input, forget, and output gates~\cite{bib25,  atkinson1971control}. These gates control the flow of information by determining what information to add, retain, or discard at each time step, enabling the network to learn and maintain dependencies over long sequences effectively. Each LSTM cell receives information from the previous cell and can pass significant long-term dependencies forward, making it well-suited for time-series data, language processing, and other sequence-based tasks~\cite{bhandari2022predicting, rimal2023comparative, bhandari2024implementation, rimal2024identifying, rimal2024real, bib25}.

In classification tasks, LSTMs are useful for problems involving temporal or sequential data, such as text classification, sentiment analysis, or speech recognition. By processing data step-by-step, LSTMs can recognize patterns in the sequence and make predictions based on the accumulated knowledge in their cells. With its distinctive architecture, the LSTM has become a core model for handling sequential data across various domains~\cite{bib26, van2020review, mienye2024recurrent, nazareth2023financial}.

\subsection{GRU}
\label{subsec8}

The GRU model, introduced by Cho et al. in 2014,  simplifies the LSTM model by removing the redundancies of maintaining both hidden state and cell state in LSTM while maintaining its effectiveness~\cite{bib27, rithani2023review}. GRUs are built with two main gates—an update gate and a reset gate—that control the flow of information, allowing the network to retain essential information over time without the more complex gating mechanisms found in LSTM. The update gate in a GRU combines the roles of the LSTM's forget and input gates, while the reset gate helps manage short-term dependencies. This streamlined architecture makes GRUs faster and computationally more efficient and often delivers improved performance on tasks with simpler sequential dependencies~\cite{bib27, rimal2023comparative, bhandari2024implementation, rimal2024identifying, rimal2024real}.

In classification problems involving sequential data, such as time series forecasting, text classification, sentiment analysis, or machine translation, GRUs are advantageous due to their simpler structure and faster training times compared to LSTMs. For instance, GRUs effectively capture temporal dependencies in time-series forecasting by focusing on the most relevant patterns within sequential numerical data. The balance between performance and efficiency has positioned GRUs as a preferred model in applications such as stock price prediction, energy demand forecasting, and sensor data analysis~\cite{bib28, bib52, shiri2023comprehensive, nazareth2023financial}.

\subsection{Transformers}
\label{subsec9}

The transformer model, introduced by Vaswani et al. in 2017, marked a breakthrough in sequence processing by overcoming the limitations of traditional models such as RNNs and LSTM networks~\cite{bib24}. Unlike RNNs, which process sequences in a step-by-step manner, the transformer model processes sequences in parallel, allowing it to capture complex dependencies and long-range relationships with greater computational efficiency~\cite{bib39}.

Transformers, originally designed for natural language processing (NLP), are increasingly gaining attention in other domains due to their exceptional ability to handle sequential data and time-series analysis. Their self-attention mechanism  makes them well-suited for capturing complex patterns in temporal data. As highlighted by Zhou et al. (2023), transformers show significant promise in long sequence time-series forecasting (LSTF), addressing challenges such as scalability and computational efficiency. By leveraging mechanisms like sparse attention and efficient decoders, transformers can handle the computational demands of long sequences while maintaining superior predictive performance, making them valuable tools for a variety of time-series applications~\cite{bib41, bib42}.

\begin{figure}[H] % force exact placement
    \centering
    \includegraphics[width= 0.9\textwidth]{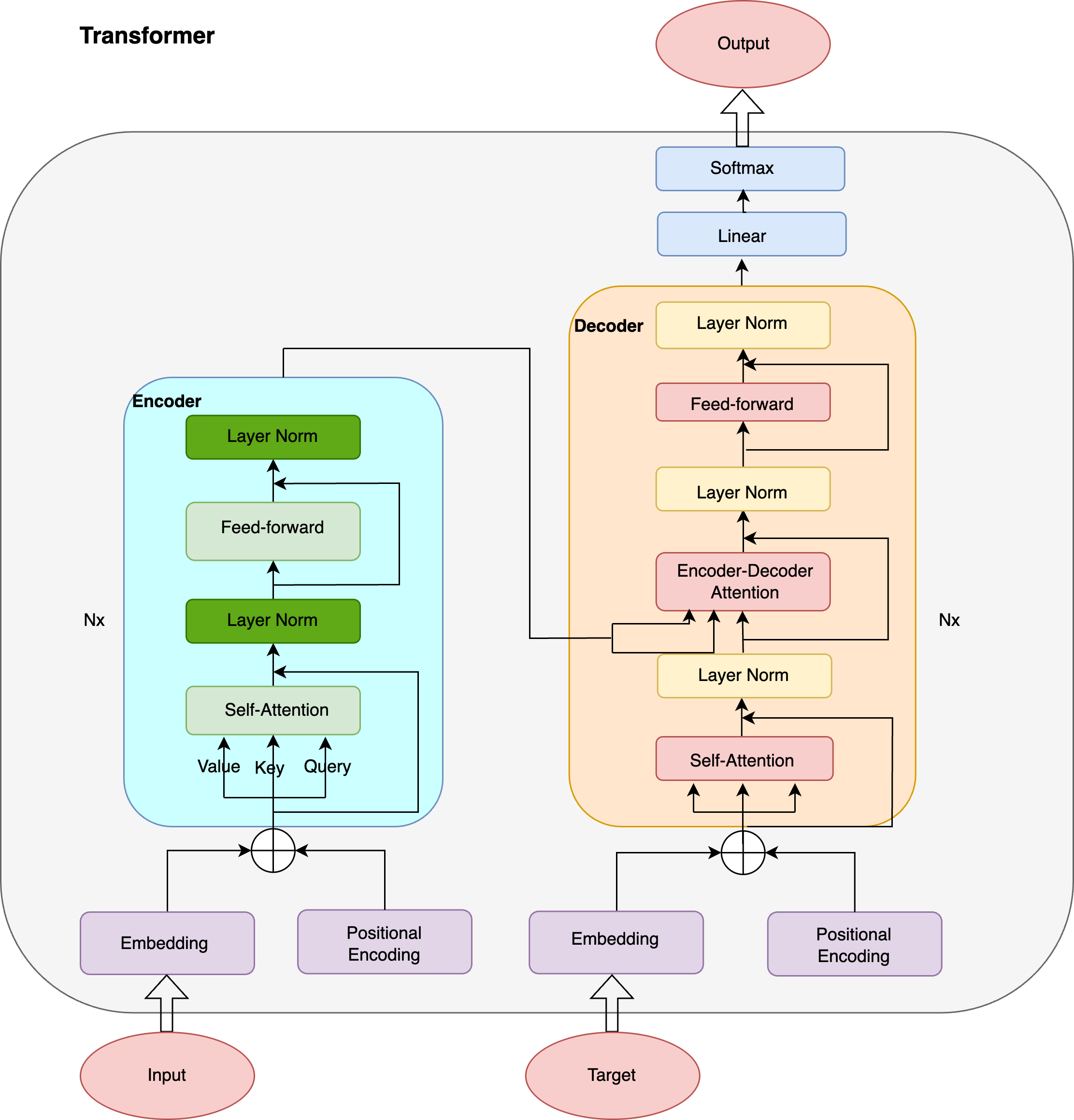}
    \caption{Transformer model architecture following the original article~\cite{bib24}}
    \label{fig:Transformer}
\end{figure}

The architecture of the transformer is built on two main components: an encoder and a decoder, as illustrated in Figure~\ref{fig:Transformer}. Each component comprises a stack of layers that apply a self-attention mechanism and a feed-forward network. The encoder generates contextualized representations of the input sequence, while the decoder uses these representations to generate an output sequence. This encoder-decoder structure is particularly advantageous for tasks such as translation, where one sequence must be transformed into another. However, for tasks requiring a single output, such as classification, the encoder alone can be utilized, as in our study~\cite{bib24, bib40}.

\subsubsection{Key Components of the Transformer Model}

\begin{description}

    \item[\normalfont\textbf{Embedding:}] In the transformer architecture, input embeddings convert input tokens—such as words, subwords, or characters—into dense, continuous vectors that capture their semantic meaning. This step is essential because transformers operate on continuous-valued data. These vectors have a fixed dimension and are trainable, enabling the model to learn meaningful token representations during training. Serving as the model's initial input, embeddings provide a compact and informative representation for subsequent layers to process.

    \item[\normalfont\textbf{Positional Encoding:}] Since the transformer processes tokens in parallel, positional encoding is added to the input embeddings to inform the model about the position of each token in the sequence. Sine and cosine functions are used to generate unique positional encodings that are added to the embeddings, allowing the model to differentiate tokens based on their positions~\cite{bib24, bib24}.
    
    \item[\normalfont\textbf{Self-Attention Mechanism}] At the core of the transformer model is the self-attention mechanism, which allows each token in a sequence to focus on other tokens based on their relevance~\cite{bib24, bib43}. For each token, the model computes three vectors: a Query (Q), a Key (K), and a Value (V). These vectors determine how much "attention" each token should pay to others in the sequence~\cite{bib44}. The self-attention score is calculated as:
    
    \[
    \text{Attention}(Q, K, V) = \text{softmax}\left(\frac{Q K^T}{\sqrt{d_k}}\right) V
    \]
    
    Here, \( d_k \) represents the dimension of the Key vector, scaling the dot product to maintain numerical stability. The softmax function normalizes the attention scores, yielding a probability distribution where each token "attends" to others in proportion to their relevance.

    \item[\normalfont\textbf{Multi-Head Attention:}] To capture a variety of relationships within the data, the self-attention mechanism is applied across multiple attention "heads." Each head learns distinct patterns and dependencies, and the results from all heads are concatenated and linearly transformed, enriching the model’s capacity to understand diverse interactions~\cite{bib24}.

    \item[\normalfont\textbf{Feed-Forward Network (FFN):}] Each token representation is passed through a feed-forward network consisting of two linear transformations with a ReLU (Rectified Linear Unit) activation function in between. This network introduces non-linearity, enabling the model to capture more complex patterns in the data.

    \item[\normalfont\textbf{Layer Normalization and Residual Connections:}] Layer normalization is applied after the self-attention and feed-forward components to stabilize training, and residual connections help prevent gradient vanishing, allowing for deeper learning. This structure is repeated through multiple encoder layers, usually six or more, to progressively refine the representations of the input sequence.
\end{description}

\subsubsection{Adaptation of the Transformer for Classification Tasks}

In our study, the original transformer model was modified and adapted to perform classification on physiological data collected from wearable sensors during exams. Specifically, only the encoder component of the transformer was retained, as our task requires generating a single output—predicted performance category—rather than producing a sequential output. The details are given below:

\begin{figure}[H] % force exact placement
    \centering
    \includegraphics[width= 0.5\textwidth]{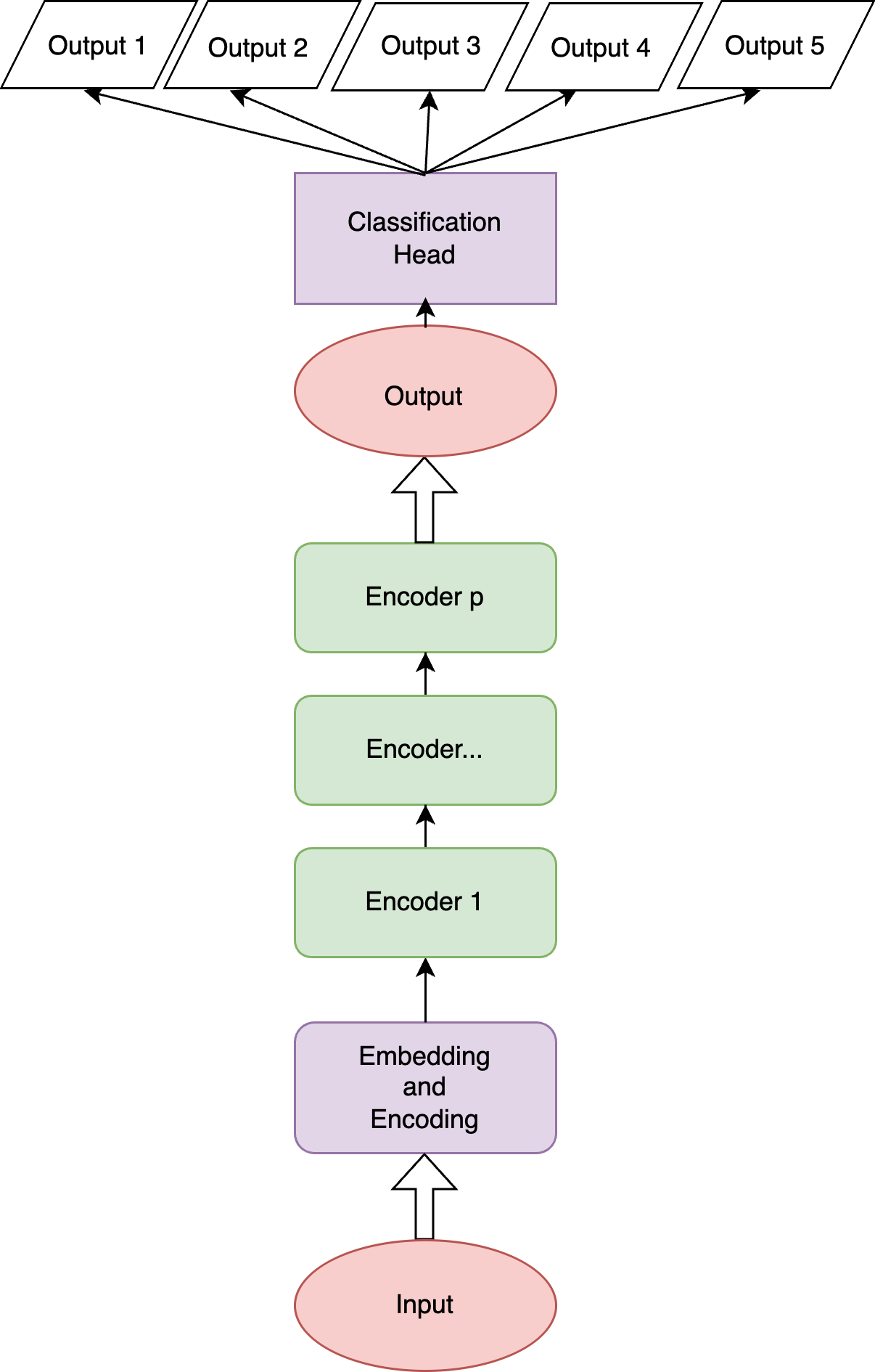}
    \caption{Transformer with the classification head}
    \label{fig:Class_h}
\end{figure}

%\subsubsection*{Modifications for Classification}

\begin{description}
    \item[\normalfont\textbf{Removal of the Decoder:}] Since this study focuses on classification instead of sequence-to-sequence tasks, the decoder component of the transformer was removed. The encoder alone is sufficient to create a contextualized representation of the input sequence, which can then be directly utilized for classification~\cite{gao2022encoder,fu2023decoder} .

    \item[\normalfont\textbf{Addition of a Classification Head:}] To adapt the encoder’s output for classification, a classification head was added. This component consists of a linear transformation followed by a softmax function, which converts the encoded representations into probabilities for each output (grade categories A, B, C, D, and F). The final predicted category corresponds to the class with the highest probability, as illustrated in Figure~\ref{fig:Class_h}.
\end{description}

\subsubsection{Application to Physiological Data Classification}

In this study, each physiological feature is treated as an individual token within the input sequence. Each sequence corresponds to a student’s physiological data collected during an exam, providing a multivariate perspective of their responses to academic stress.

\begin{description}
    \item[\normalfont\textbf{Embedding and Positional Encoding:}] The input physiological data is first embedded, with each feature represented as a dense vector capturing its key characteristics. Positional encoding is added to these embeddings to maintain the order of the sequence, allowing the model to recognize the temporal structure of the physiological responses.

    \item[\normalfont\textbf{Self-Attention and Multi-Head Attention:}] The embedded input passes through the multi-head self-attention mechanism, where the model evaluates interactions among physiological features over the sequence. This allows the transformer to capture both short- and long-range dependencies between physiological features, which is crucial for understanding how different aspects of stress response contribute to cognitive performance.

    \item[\normalfont\textbf{Final Prediction Model:}] After processing through multiple encoder layers, the output of the encoder stack represents a contextualized view of the entire input sequence. This output is passed to the classification head, which assigns probabilities to each performance category. The model’s final output is the predicted grade category with the highest probability, reflecting the model’s interpretation of the physiological responses in relation to academic performance.
\end{description}

By adapting the transformer model in this way, our study leverages the model’s strengths in handling complex dependencies within multivariate data, providing a powerful framework for predicting cognitive performance based on physiological responses during exams. This approach allows us to capture a comprehensive picture of each individual’s physiological response, offering insights into the potential impact of stress on academic outcomes. To the best of our knowledge, this is the first attempt to implement transformer architecture to model academic outcomes using physiological signals. 

\section{Evaluation Framework and Results}\label{sec5}

This study focuses on performing a comparative analysis of machine learning models for predicting academic outcomes. To achieve the research objectives, the experimental process is systematically designed in six phases: (1) preprocessing the physiological data and preparing it for analysis, (2) constructing model architectures and performing hyperparameter tuning, (3) identifying the best-performing models within each algorithmic category, (4) comparing models across categories to evaluate their predictive power, (5) conducting statistical analyses to validate the results, and (6) identifying the overall best model based on a comprehensive evaluation. Each phase is carefully structured to ensure a reproducible experimental design.

\subsection{Experimental Setup}\label{subsec10}

The study employed Python as the primary programming language, utilizing Scikit-learn to implement and evaluate metrics for traditional machine learning algorithms, including logistic regression, SVM, and random forest. TensorFlow (via Keras APIs) was leveraged to develop and assess deep learning models such as GRU and LSTM. Additionally, the transformer model was adapted from code available in various GitHub repositories [\href{https://github.com/hyunwoongko/transformer}{1} \href{https://github.com/mselmangokmen/TimeSeriesProject}{2}].    PyTorch was used to build and evaluate the transformer models. Performance metrics—namely accuracy, precision, recall, and F1 Score—were utilized to compare the model performance. The experiments were conducted on a Dell Precision 3660 Tower equipped with a 13th generation Intel i9-1300k 3.00GHz processor, 128 GB of RAM, and an Nvidia GeForce RTX 4090 with 24GB GDDR6.

For data preparation, the dataset was divided into training, validation, and test sets. The training set comprised 85\% of the data, while the remaining 15\% was designated for testing, as illustrated in Table~\ref{Table: data_split}. Additionally, 15\% of the training set was set aside as a validation set during hyperparameter tuning to optimize the models. The experimented models use physiological and categorical input features to predict the exam outcomes.

%%%%%%%%%%%%%%%%%%%
\begin{table}[H]
\centering  
\caption{Data split for training and test sets}\label{Table: data_split}
\begin{tabular}{c|c|c}
\textbf{Data} & \textbf{No. of Samples} & \textbf{Proportion (\%)} \\ 
\hline
Complete data & 328,655 & 100 \\ 
Training and Validation & 279,356 & 85 \\ 
Test & 49,299 & 15 \\ 
\end{tabular}
\end{table}
%%%%%%%%%%%%%%%%%%%

\subsection{Model Development and Hyperparamaeter  Optimization}\label{subsec11}

Hyperparameter tuning involves selecting the best combination of hyperparameters for a machine learning model to optimize its performance on unseen data. Unlike parameters learned during training, hyperparameters are set before the training process, significantly impacting the model's generalization ability. Effective hyperparameter tuning is crucial for balancing underfitting and overfitting, which improves the model's predictive accuracy and robustness. This study carefully examines hyperparameter tuning for the models used, and various models were evaluated using distinct hyperparameter tuning strategies tailored to their characteristics:  

\textbf{Deep Learning Models (Transformers, LSTM, GRU):}  These models were trained for a maximum of 100 epochs per run, utilizing early stopping based on validation loss with patience of five epochs, to  train for sufficient  epochs while  preventing overfitting. Cross-validation was not used due to the computational expense of such complex architectures.  The parameters for the LSTM and GRU were tuned using Keras Tuner, with the number of layers ranging from 1 to 4, the number of neurons per layer ranging from 16 to 512, dropout rates between layers ranging from 0.0 to 0.3, and optimizers chosen from Adam, Nadam, and AdamW. The optimizers used exponential decay with an initial learning rate of 0.001, a decay rate of 0.95, and staircase set to true. For the transformer model, the number of heads ranged between 2 and 16, the number of layers ranged between 2 and 6, dropout probabilities ranged between 0.0 and 0.3, the dimension for positional embedding ($d\_model$) was either 256 or 512, the number of neurons in the feed forward network ranged between 64 and 256, batch sizes ranged between 64 and 256, and sequence lengths ranged between 64 and 256.  

\textbf{Logistic Regression:}  GridSearchCV using 5-fold cross-validation was used to tune hyperparameters with the \texttt{multi\_class = multinomial} setting. The grid included the regularization strength parameter $C$ ranging from 0.001 to 100, regularization types (l1, l2, and elasticnet), and different solver options. The maximum iteration limit was set to 10000 to ensure convergence during optimization.  

\textbf{SVM:}  
GridSearchCV using 5-fold cross-validation was used to tune hyperparameters, including the regularization strength $C$, kernel types, and kernel-specific parameters. The penalty parameter ($C$) was chosen in the range of 0.1 to 100. For the Gaussian kernel, $\gamma$ values ranging from 0.01 to 1 were tested. For the polynomial kernel, degrees up to 5 and independent term coefficients ($coef0$) between 0 and 3 were explored. 

\textbf{Random Forest:}  GridSearchCV was employed to tune parameters such as the number of estimators ($n\_estimators$ ranging between 100 and 200), maximum tree depth ($max\_depth$ between 5 and 15), and the minimum number of samples required to split a node ($min\_samples\_split$ between 5 and 50), using 5-fold cross-validation.  

The best list of hyperparameters is chosen based on average accuracy scores from ten replicates on the validation set for deep learning model configurations and the average cross-validation score on the other models. Table~\ref{tab:hyperparameters} reports the best set of hyperparameters learned in our experimental setting discussed above. The dropout rates are provided as a list of the rates applied after each layer in the order respectively.

\begin{table}[h!]
\centering
\caption{Machine Learning Models and Their Tuned Parameters}
\label{tab:hyperparameters}
\begin{tabular}{|l|l|}
\hline
\textbf{Model}                  & \textbf{Parameters}                                                                                 \\ \hline
\textbf{Transformer}            & Number of heads=8, Number of layers=3, Dropout probability=0, \\ 
                                 & $d\_model =256$, Hidden layer neurons= 128, Batch size = 128, \\ 
                                 & Sequence length = 128.                                                                                  \\ \hline
\textbf{LSTM}                   & Number of layers = 4, Neurons in layer= [512, 256, 512, 256], \\ 
                                 & Optimizer= Adam, Learning rate= 0.001, Decay rate= 0.95, \\ 
                                 & Batch size = 256, Dropout rate=[0.0, 0.1, 0.3, 0.2].                                                                                       \\ \hline
\textbf{GRU}                    & Number of layers = 3, Neurons in layer=[256, 256,128], \\ 
                                 & Optimizer= AdamW, Learning rate=0.001, Decay rate=0.95, \\ 
                                 & Batch size =256  , Dropout rate=[ 0.0, 0.1, 0.3] .                                                                                   \\ \hline
\textbf{Logistic Regression}    & Regularization strength ($C= 1$), Regularization type= l1, \\ 
                                 &  Solver = saga, Maximum iterations = 10000.                                                                            \\ \hline
\textbf{SVM} & Regularization strength ($C=10$), Kernel type= rbf, $\gamma=0.01$.  \\ \hline
\textbf{Random Forest}          & Number of estimators =200, Maximum depth= 15, \\ 
                                 & Minimum samples for split =5.                                                \\ \hline
\end{tabular}
\end{table}

Once the hyperparameters were tuned, we proceeded to train each configuration using the complete training dataset. To ensure a fair and robust comparison of model performance, each configuration was replicated 30 times, and the average performance on the test data was reported. The deterministic nature of SVM and logistic regression results in identical performance across training runs when using the same datasets and hyperparameters, leading to no variability in evaluation metrics. In contrast, due to the randomness inherent in building trees in random forests and the stochastic behaviors observed during model initialization in deep learning methods, there is variability in model performance across different runs. We omitted logistic regression and SVM models for the variability analysis due to their deterministic characteristics in this context.

\subsection{Model Evaluation and Performance Comparison }\label{subsec12}

\begin{figure}[H] % force exact placement
    \centering
    \includegraphics[width= 1.0\textwidth]{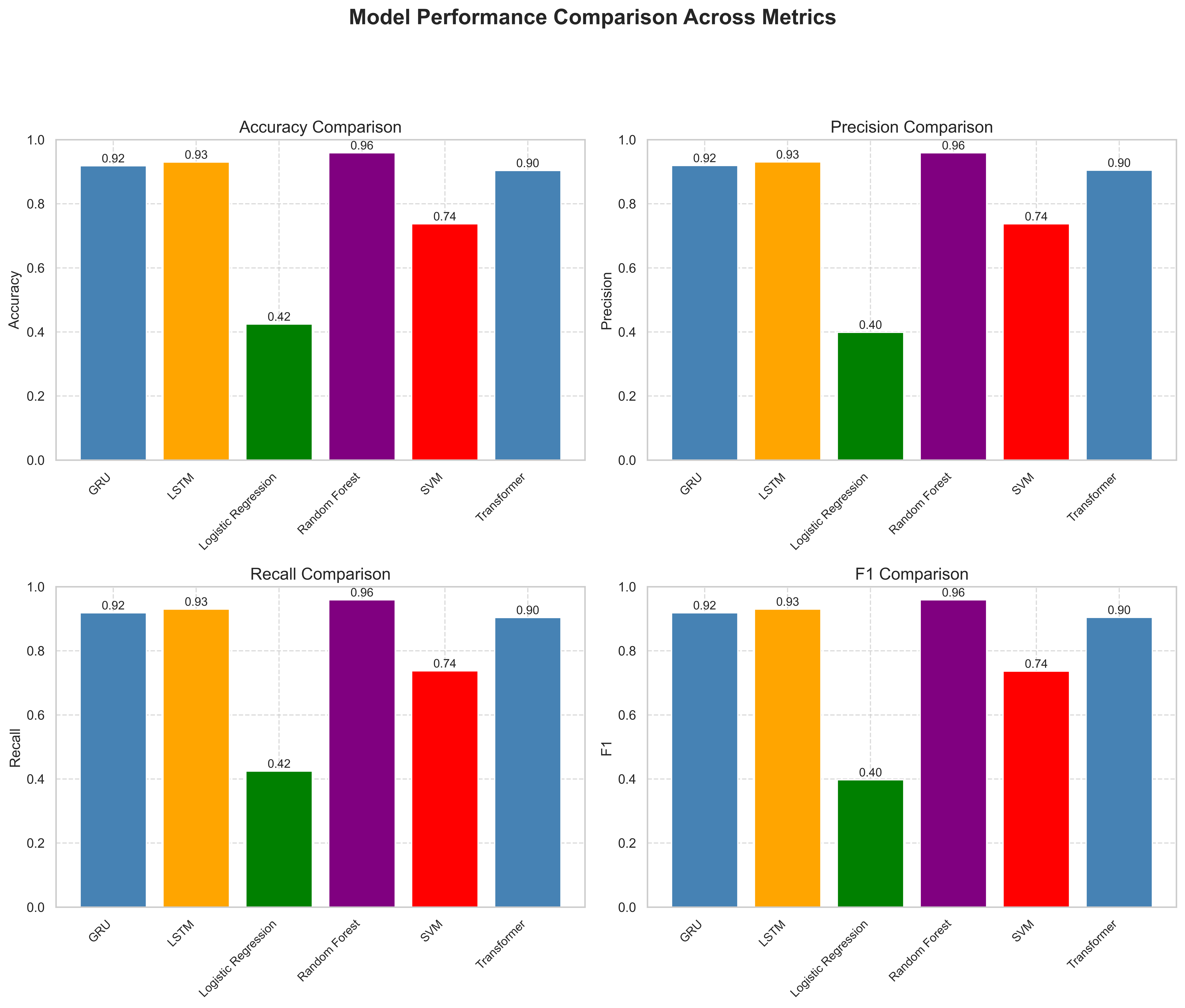}
    \caption{Bar plots comparing model performance across evaluation metrics: top-left (accuracy), top-right (precision), bottom-left (recall), and bottom-right (F1 score)}
    \label{fig:Model_perf}
\end{figure}

\begin{figure}[H] % force exact placement
    \centering
    \includegraphics[width= 1.0\textwidth]{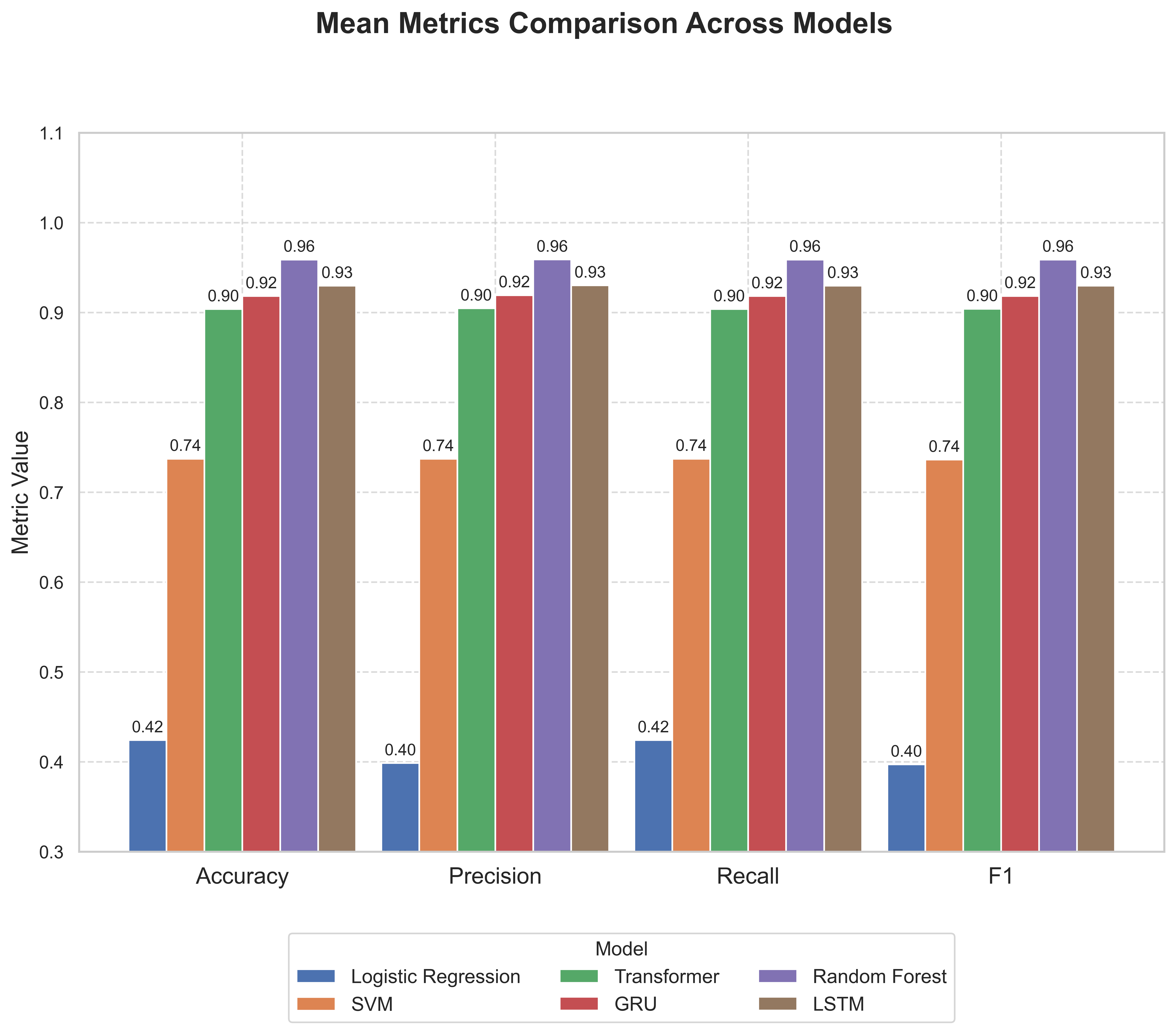}
    \caption{Comparison plot of mean performance metrics across replications for all models }
    \label{fig:Metrics_Comp}
\end{figure}

The models are evaluated on the test data, and their performance is presented through tables and visuals. The optimal model is selected based on the average performance of each model configuration on the test data.
For this, we examine the model performance comparison across metrics (Figure~\ref{fig:Model_perf}) and the mean metrics comparison across models (Figure~\ref{fig:Metrics_Comp}). Together, these visualizations offer a comprehensive understanding of the mean performance and consistency of the models.

The comparison of model performance, illustrated in Figure~\ref{fig:Model_perf} and Figure~\ref{fig:Metrics_Comp}, reveals distinct strengths and limitations among the different approaches. The GRU and LSTM models yield reliable results, with mean performance metrics of 0.92 and 0.93, respectively, demonstrating excellent generalization to unseen data. The transformer model has a slightly lower mean metric of approximately 0.90, but it shows versatility by obtaining dependable performance in modeling the tabular data. In contrast, the random forest model achieves the highest mean metric of 0.96, effectively modeling non-linear relationships and proving the best choice for this particular modeling problem. Meanwhile, SVM exhibit moderate performance with mean metrics around 0.74. Logistic regression demonstrates the lowest performance with mean metrics below 0.42, making it ill-suited for handling the dataset's complexity.

\vspace{-0.1 in}
\begin{figure}[H] % force exact placement
    \centering
    \includegraphics[width= 1\textwidth]{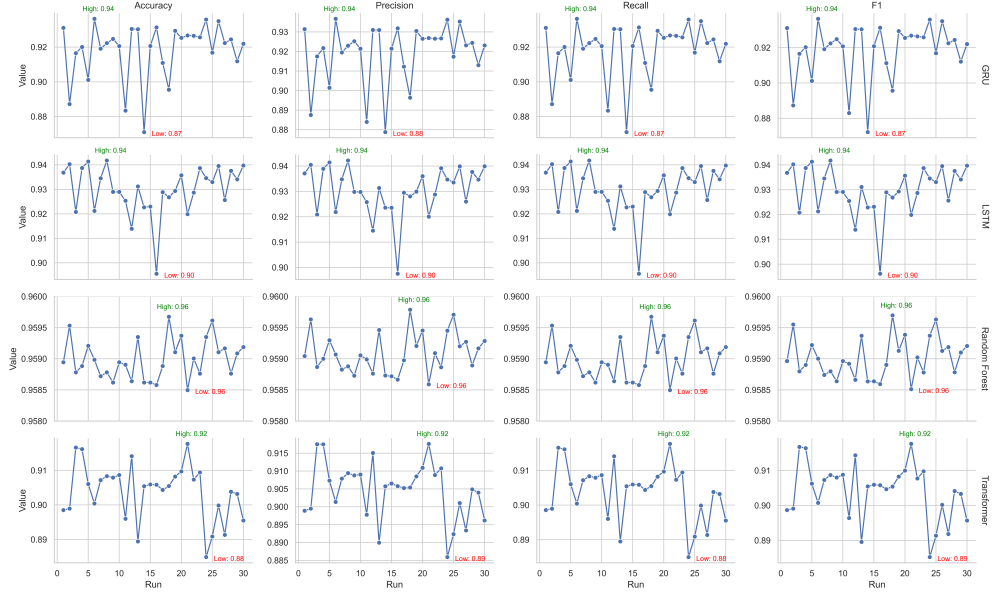}
    \caption{Trend analysis of model performance across 30 replicates: GRU (top row), LSTM (second row), random forest (third row), and transformer (bottom row)}
    \label{fig:trend_analysis}
\end{figure}

\vspace{-0.1 in}
\begin{figure}[H] % force exact placement
    \centering
    \includegraphics[width= 1.0\textwidth]{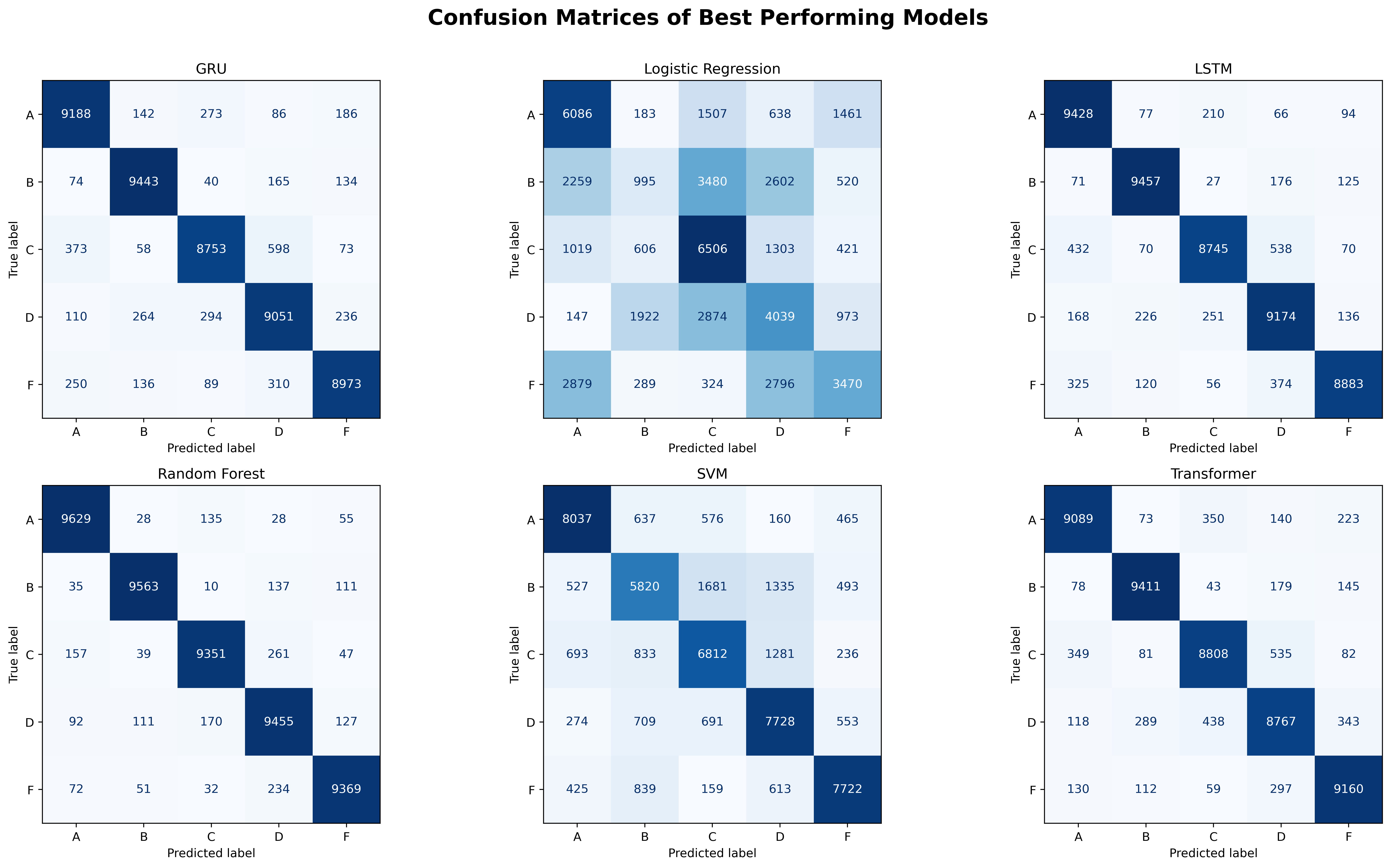}
    \caption{Confusion matrices of the best-performing model replicates—GRU (top left), logistic regression (top middle), LSTM (top right), random forest (bottom left), SVM (bottom middle), and transformer (bottom right)—showing the number of correct and incorrect predictions for each class}
    \label{fig:all_confusion_matrices}
\end{figure}

\begin{figure}[H] % force exact placement
    \centering
    \includegraphics[width= 1.02\textwidth]{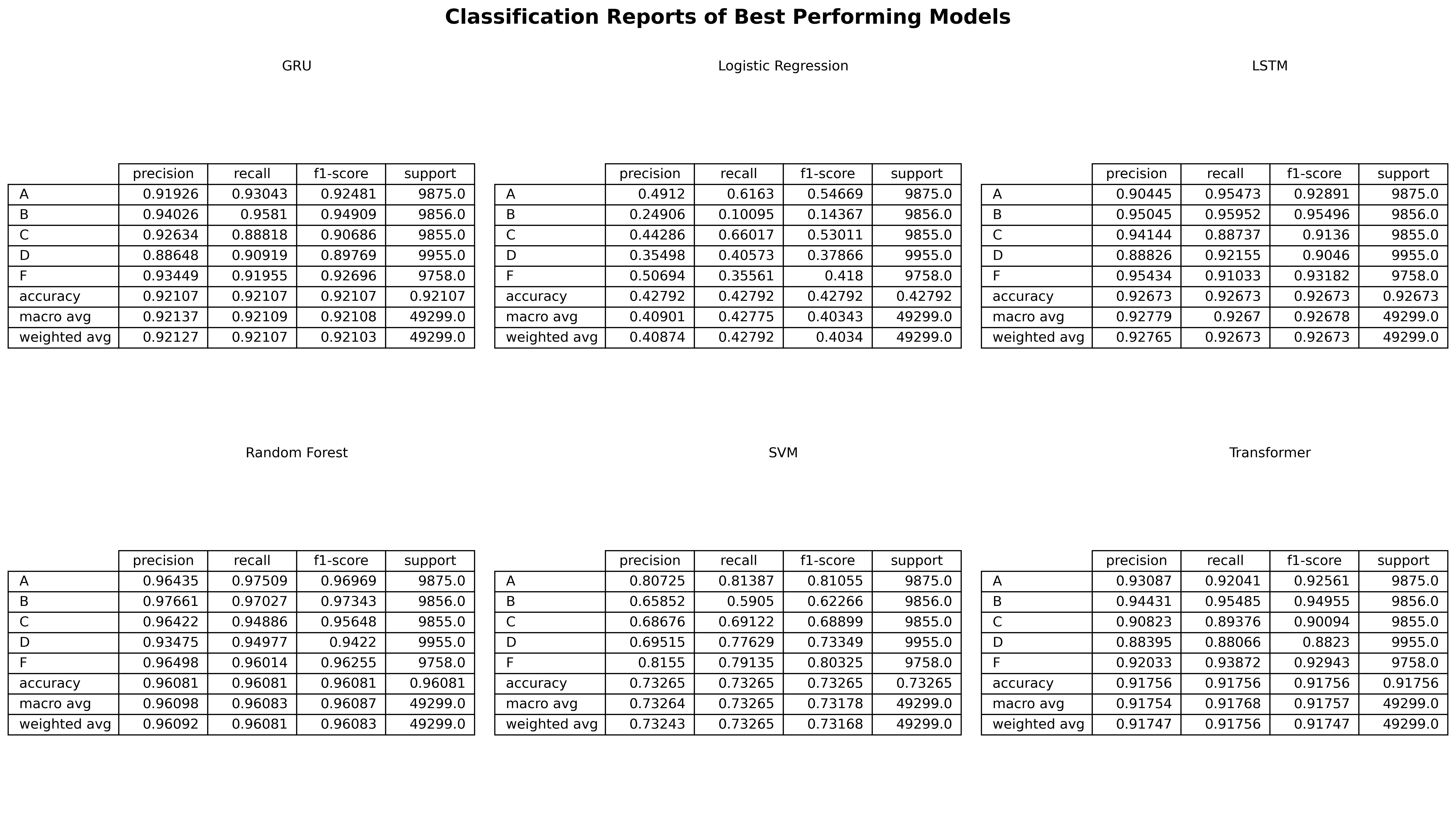}
    \caption{Classification reports of the best-performing model replicates: GRU (top left), logistic regression (top middle), LSTM (top right), random forest (bottom left), SVM (bottom middle), and transformer (bottom right) illustrating variations in performance across metrics and classes}
    \label{fig:all_classification_reports}
\end{figure}

\begin{table}[h]
    \centering

    \begin{tabular}{l|c|c|c|c}
        \hline
        \textbf{Model} & \textbf{Metric} & \textbf{Mean $\pm$ Std} & \textbf{Min} & \textbf{Max} \\
        \hline
        Logistic Regression & Accuracy & 0.4243 $\pm$ 0.0000 & 0.4204 & 0.4282 \\
                            & Precision & 0.3988 $\pm$ 0.0000 & 0.3936 & 0.4040 \\
                            & Recall & 0.4243 $\pm$ 0.0000 & 0.4204 & 0.4282 \\
                            & F1 Score & 0.3973 $\pm$ 0.0000 & 0.3927 & 0.4025 \\
        \hline
        SVM & Accuracy & 0.7374 $\pm$ 0.0000 & 0.7374 & 0.7374 \\
            & Precision & 0.7372 $\pm$ 0.0000 & 0.7372 & 0.7372 \\
            & Recall & 0.7374 $\pm$ 0.0000 & 0.7374 & 0.7374 \\
            & F1 Score & 0.7364 $\pm$ 0.0000 & 0.7364 & 0.7364 \\
        \hline
        Transformer & Accuracy & 0.9039 $\pm$ 0.0081 & 0.8849 & 0.9177 \\
                    & Precision & 0.9048 $\pm$ 0.0080 & 0.8859 & 0.9176 \\
                    & Recall & 0.9039 $\pm$ 0.0081 & 0.8849 & 0.9177 \\
                    & F1 Score & 0.9041 $\pm$ 0.0080 & 0.8852 & 0.9176 \\
        \hline
        GRU & Accuracy & 0.9185 $\pm$ 0.0159 & 0.8711 & 0.9363 \\
            & Precision & 0.9194 $\pm$ 0.0152 & 0.8788 & 0.9367 \\
            & Recall & 0.9185 $\pm$ 0.0159 & 0.8711 & 0.9363 \\
            & F1 Score & 0.9186 $\pm$ 0.0159 & 0.8722 & 0.9363 \\
        \hline
        LSTM & Accuracy & 0.9299 $\pm$ 0.0098 & 0.8955 & 0.9418 \\
             & Precision & 0.9304 $\pm$ 0.0095 & 0.8975 & 0.9422 \\
             & Recall & 0.9299 $\pm$ 0.0098 & 0.8955 & 0.9418 \\
             & F1 Score & 0.9300 $\pm$ 0.0097 & 0.8961 & 0.9418 \\
        \hline
        Random Forest & Accuracy & 0.9590 $\pm$ 0.0003 & 0.9585 & 0.9597 \\
                      & Precision & 0.9591 $\pm$ 0.0003 & 0.9586 & 0.9598 \\
                      & Recall & 0.9590 $\pm$ 0.0003 & 0.9585 & 0.9597 \\
                      & F1 Score & 0.9590 $\pm$ 0.0003 & 0.9585 & 0.9597 \\
        \hline
    \end{tabular}
    \caption{Performance statistics of all the machine learning models on test data}
    \label{tab:performance_stats}
\end{table}

Figure~\ref{fig:trend_analysis} and Table~\ref{tab:performance_stats} illustrate the model's consistency, variability, and average scores in all the performance metrics over 30 experimental runs. Given that the SVM and logistic regression models do not exhibit stochastic behavior, their respective analyses are not included in the figure.   Interestingly, the random forest model outperformed all others, achieving the highest mean accuracy of 0.9590 with the smallest variability(0.0003), underscoring its consistency, precision,  and computational efficiency. Moreover, the model's inferential power positions it as the optimal choice for our problem domain. The LSTM models consistently achieved high-performance metrics with a mean accuracy of 0.9299 and variability of less than one percent, demonstrating their robustness and strong generalization capabilities. The GRU models also consistently achieved high-performance metrics with a mean accuracy of 0.9185  and a standard deviation of 0.0159. While exhibiting an average accuracy of 0.9039  and a standard deviation of 0.0081, slightly lower than the accuracy of GRU and LSTM, the transformer model maintained reliable and consistent results, indicating its adaptability and effectiveness for multivariate classification tasks. In contrast, the SVM model achieved a moderate accuracy of 0.7374, while the logistic regression model exhibited poor performance with a mean accuracy of 0.4243, thereby struggling to adequately capture the non-linear and multivariate complexities inherent in the dataset. The other performance metrics followed a similar pattern to accuracy for all the models compared.

Furthermore, Figures~\ref{fig:all_confusion_matrices} and \ref{fig:all_classification_reports} illustrate the confusion matrix and classification report for the best-performing replicates of all the models analyzed. There is a noticeable variation in each model's performance across different classes. Among these, the logistic regression model exhibits the highest variability in performance scores for all classes across all metrics, followed by SVM being the second highest. The LSTM, GRU, and transformer models also demonstrate overall low and comparable variability in performance metrics across all classes. The random forest model exhibits the lowest variation in performance across different classes, demonstrating the best model overall with the highest overall performance scores and lowest variability between performance across classes and also across runs.

\subsection{Statistical Analysis}\label{subsec_STA}
To evaluate the performance differences among the predictive models, we conducted a statistical analysis based on results from 30 independent runs of each model on the test set. The SVM model was excluded from this analysis due to its deterministic outputs, which showed no variation across runs. The models included in the analysis were GRU, LSTM, logistic regression, transformer, and random forest. Before performing statistical tests, we assessed the distribution of each model's performance metrics using the D'Agostino and Pearson omnibus test for normality. The GRU and LSTM exhibited deviations from normality, so we applied both parametric and nonparametric tests to ensure comprehensive comparisons. Pairwise statistical comparisons were conducted using two complementary tests. Welch's t-test is a parametric method chosen for its ability to handle unequal variances between groups. In addition, the Mann–Whitney U test, a nonparametric alternative, was used to account for cases where the data did not follow a normal distribution.

We previously observed that random forest consistently outperformed all other models across all evaluation metrics. The t-tests and Mann–Whitney U tests produced p-values below the 0.05 significance level in all pairwise comparisons, confirming that the observed performance differences were statistically significant and not likely due to chance. These results further support the superiority of the random forest model for this classification task. The corresponding QQ plots and Tables summarizing the statistical test results are in the Appendix.

\subsection{Final Model Selection and Evaluation}\label{subsec14}

\begin{figure}[H] % force exact placement
    \centering
    \includegraphics[width= 0.8\textwidth]{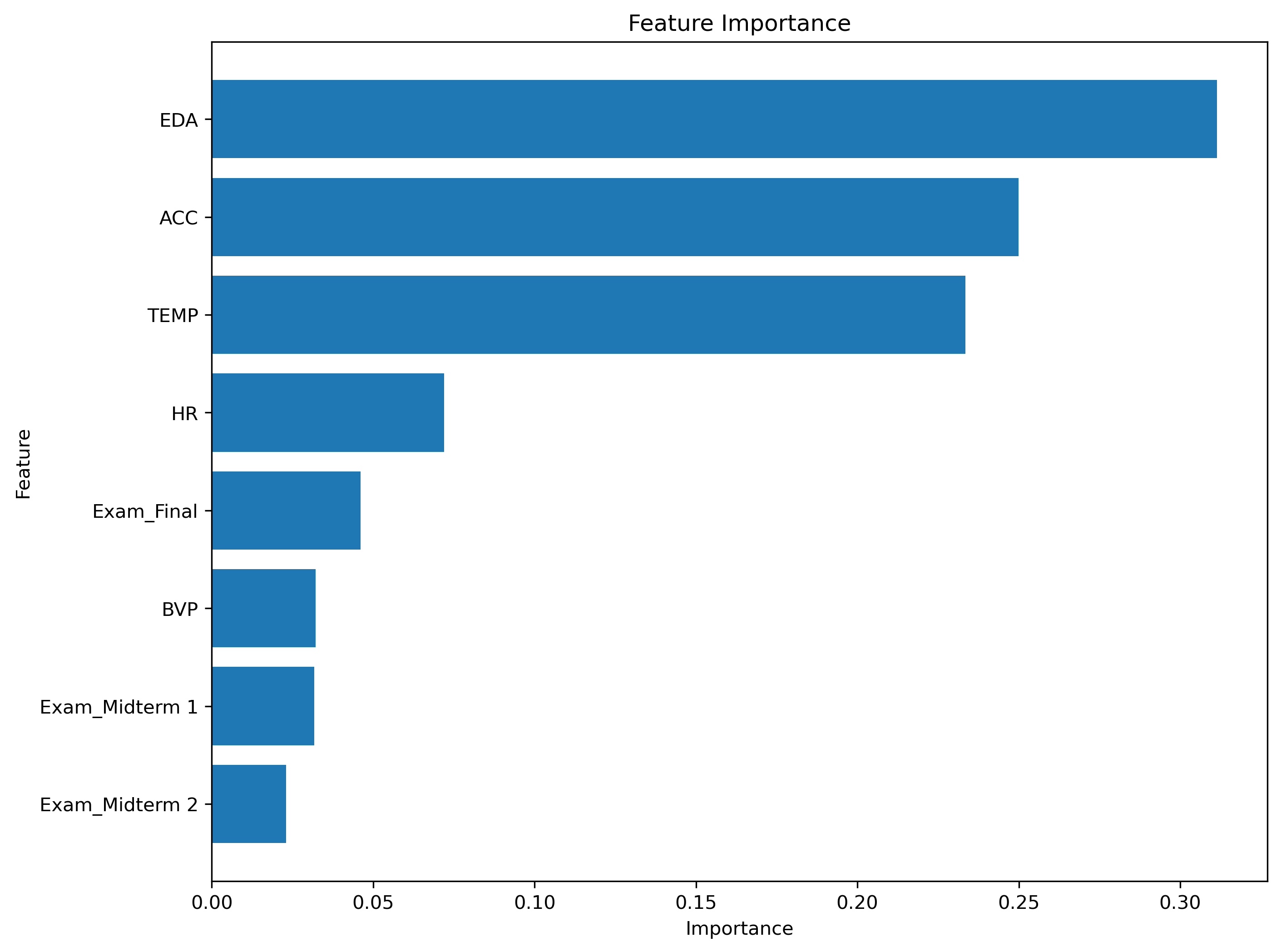}
    \caption{Feature importance plot of the random forest model, where longer bars indicate features with greater contribution to the model’s predictions}
    \label{fig:feature_import}
\end{figure}

This section synthesizes insights from previous analysis to identify a model that excels across all performance metrics on unseen data while maintaining computational efficiency and robustness. For this, we compare the strengths and limitations of deep learning and traditional machine learning approaches, identify the best model in each class, and then identify the overall best model. Among the deep learning models evaluated, the LSTM achieved the highest mean accuracy of 0.9299 with a low standard deviation of 0.0098, indicating reliability and robustness. The GRU closely followed at 0.9185, demonstrating robustness despite slightly higher variability. The transformer model yielded a mean accuracy of 0.9039 and a standard deviation of 0.0081, demonstrating a promising model in handling the regular tabular data with its self-attention mechanism, though it slightly underperformed compared to LSTM and GRU. The transformer's ability to generalize effectively underscores its potential for broader applications. These qualities make it a forward-looking model, aligning closely with the study's objective to evaluate its adaptability and efficacy in handling numerical data for classification tasks, particularly in structured tabular data.

Among the standard machine learning models, the random forest reached the highest mean accuracy of 0.9590 with an exceptionally low standard deviation of 0.0003, underscoring its predictive consistency and efficiency. Its feature importance analysis revealed EDA, ACC, and TEMP as the top three predictors for exam outcomes, providing additional insights for practical application in cognitive performance assessment. Conversely, logistic regression recorded a mean accuracy of 0.4243, unable to capture the dataset's complex interactions. The variable importance of the logistic regression model was not discussed due to its low performance across all metrics, making it unreliable. SVM achieved a moderate accuracy of 0.7374; however, since the infinite-dimensional kernel was used, obtaining the actual feature importance was not feasible. The statistical analysis supports the superiority in performance of the random forest; it is statistically significantly better than other models.

In summary, while deep learning models also showed strong performance across all metrics and demonstrated consistency, the random forest model stood out as the most promising option. It effectively balances robust performance, efficiency, and interpretability, making it well-suited for modeling the physiological data used in this study. Additionally, the inferential insights gained from its analysis allow for exploration of the potential implications of the top features in other high-stress work environments.

\section{Ethics and Practical Implications}\label{sec6}

This study upholds ethical integrity throughout its design, implementation, and reporting process, ensuring transparency and accountability. The use of publicly accessible datasets and fully documented computational methodologies, including model parameters and hyperparameters, promotes reproducibility and reliability. To ensure trustworthy results, the study averages findings over multiple runs and reports metrics comprehensively.

While the insights gained from this study are valuable, they are not intended to stand alone as decision-making tools. Predictions of academic performance derived from physiological data should be interpreted within the context of other influential factors, such as personal circumstances, environmental conditions, and mental health considerations, to guarantee fairness and mitigate biases. High-stakes decisions, such as those related to educational interventions, must incorporate these insights within a broader context to avoid imposing undue stress and unintended consequences on students. Educators, mental health professionals, and stakeholders can leverage the framework of this study to create tools for monitoring stress and predicting performance that uphold ethical standards while promoting student well-being and mental health awareness.

 Moreover, the adaptability of this framework extends beyond academic settings, offering potential applications in areas like workplace productivity and mental health monitoring. By integrating physiological data with mental health assessments, it could facilitate the early detection of stress-related disorders and interventions aimed at enhancing overall well-being. This study underscores the importance of embedding ethical considerations into the utilization of physiological data, thereby ensuring the development of supportive tools that prioritize transparency, fairness, and participant welfare.

\section{Conclusion}\label{sec7}

Physiological signals exhibit complex temporal dynamics, multivariate dependencies, and non-linear patterns, making their analysis a challenging task that demands advanced modeling techniques. This study addresses these challenges by performing a comparative analysis of two classes of machine learning models—traditional and deep learning—using key classification metrics: accuracy, precision, recall, and F1-score. The objective was to assess the predictive capabilities of various models in classifying physiological data for cognitive performance and stress detection, and identify the influential factors.

Eight carefully curated features were extracted from the physiological data, and extensive model configurations were tested across diverse architectures and hyperparameters. Rigorous model selection, based on performance metrics, identified random forest as the most reliable model, offering consistent and interpretable results through feature importance analysis. Although random forest demonstrated superior performance, deep learning models, LSTM, GRU, and transformers consistently achieved accuracy above 90\%, underscoring their suitability for complex data. Notably, the transformer model, tested for the first time on physiological data in this study, showed very competitive results despite being slightly outperformed by LSTM, GRU, and random forest. Known as the state of the art in many problem domains, its versatility in capturing long-range dependencies and multivariate relationships highlights its potential for future applications in this domain. From the inference perspective, our experimental results demonstrated that ACC,  EDA, and TEMP are the top three features influencing academic performance.

In conclusion, this article analyzes the performance of machine learning models for physiological data classification, providing valuable insights into cognitive performance prediction and stress monitoring. Future research may explore hybrid models that integrate the strengths of different architectures to improve robustness and scalability. Expanding the dataset diversity, refining feature extraction techniques, and optimizing model efficiency will further enhance the applicability of these models in mental health monitoring and educational support systems.

%\bmhead{Acknowledgements}

{\small
% (This section intentionally left blank for now.)
}

\section*{Declarations}

\begin{itemize}
    \item \textbf{Funding:} {\small No funding was received to assist with the preparation of this manuscript.}
    \item \textbf{Conflict of interest/Competing interests:} The authors have no conflict of interest to declare that are relevant to the content of this article.
    \item \textbf{Ethics approval and consent to participate:} Not Applicable.
    \item \textbf{Consent for publication:} Not Applicable.
    \item \textbf{Data availability:} {\small All data used in this study are derived from publicly available datasets.}
    \item \textbf{Materials availability:} Not Applicable.
    \item \textbf{Code availability:} {\small Codes will be made available on GitHub to the public once the manuscript is accepted for publication.}
    \item \textbf{Author contribution:} {\small R.R and L.Y conceptualized the research problem and methodology. R.R and L.Y worked on writing and executing code. L.Y and R.R prepared the figures. All authors contributed to validation of the results, writing the original draft, reviewing, and editing the manuscript.}
\end{itemize}

\bibliography{sn-bibliography}% common bib file

@INPROCEEDINGS{bib14,
  author={Rafiul Amin, Md. and Wickramasuriya, Dilranjan S. and Faghih, Rose T.},
  booktitle={2022 IEEE Healthcare Innovations and Point of Care Technologies (HI-POCT)}, 
  title={A Wearable Exam Stress Dataset for Predicting Grades using Physiological Signals}, 
  year={2022},
  volume={},
  number={},
  pages={30-36},
  doi={10.1109/HI-POCT54491.2022.9744065}}

@article{bib15,
  title={A survey of machine learning techniques in physiology based mental stress detection systems},
  author={Panicker, Suja Sreeith and Gayathri, Prakasam},
  journal={Biocybernetics and Biomedical Engineering},
  volume={39},
  number={2},
  pages={444--469},
  year={2019},
  publisher={Elsevier}
}

@article{bib16,
  title={A review of machine learning in hypertension detection and blood pressure estimation based on clinical and physiological data},
  author={Martinez-R{\'\i}os, Erick and Montesinos, Luis and Alfaro-Ponce, Mariel and Pecchia, Leandro},
  journal={Biomedical Signal Processing and Control},
  volume={68},
  pages={102813},
  year={2021},
  publisher={Elsevier}
}

@article{bib17,
  title={Applying machine learning to continuously monitored physiological data},
  author={Rush, Barret and Celi, Leo Anthony and Stone, David J},
  journal={Journal of clinical monitoring and computing},
  volume={33},
  pages={887--893},
  year={2019},
  address = {New York},
  publisher={Springer}
}

@article{bib18,
  title={A review, current challenges, and future possibilities on emotion recognition using machine learning and physiological signals},
  author={Bota, Patricia J and Wang, Chen and Fred, Ana LN and Da Silva, Hugo Pl{\'a}cido},
  journal={IEEE access},
  volume={7},
  pages={140990--141020},
  year={2019},
  publisher={IEEE}
}

@article{bib19,
  title={Machine learning methods for fear classification based on physiological features},
  author={Petrescu, Livia and Petrescu, C{\u{a}}t{\u{a}}lin and Oprea, Ana and Mitruț, Oana and Moise, Gabriela and Moldoveanu, Alin and Moldoveanu, Florica},
  journal={Sensors},
  volume={21},
  number={13},
  pages={4519},
  year={2021},
  publisher={MDPI}
}

@misc{bib20,
  title={Machine Learning and Deep Learning for Physiological Signal Analysis},
  author={Tripathy, Rajesh Kumar and Paternina, Mario Arrieta and de la O Serna, Jos{\'e} Antonio},
  journal={Frontiers in Physiology},
  volume={13},
  pages={887070},
  year={2022},
  publisher={Frontiers Media SA}
}

@article{bib21,
  title={Transformers in time-series analysis: A tutorial},
  author={Ahmed, Sabeen and Nielsen, Ian E and Tripathi, Aakash and Siddiqui, Shamoon and Ramachandran, Ravi P and Rasool, Ghulam},
  journal={Circuits, Systems, and Signal Processing},
  volume={42},
  number={12},
  pages={7433--7466},
  year={2023},
   address = {New York},
  publisher={Springer}
}

@article{bib22,
  title={Non-stationary transformers: Exploring the stationarity in time series forecasting},
  author={Liu, Yong and Wu, Haixu and Wang, Jianmin and Long, Mingsheng},
  journal={Advances in Neural Information Processing Systems},
  volume={35},
  pages={9881--9893},
  year={2022}
}

@misc{bib23,
  title={A Wearable Exam Stress Dataset for Predicting Cognitive Performance in Real-World Settings},
  author={Amin, Md Rafiul and Wickramasuriya, Dilranjan and Faghih, Rose T},
  year={2022},
  howpublished={PhysioNet},
  version={1.0.0},
  doi={10.13026/kvkb-aj90},
  url={https://doi.org/10.13026/kvkb-aj90}
}

@article{bib24,
  title={Attention is all you need},
  author={Vaswani, Ashish and Shazeer, Noam and Parmar, Niki and Uszkoreit, Jakob and Jones, Llion and Gomez, Aidan N and Kaiser, {\L}ukasz and Polosukhin, Illia},
  journal={Advances in neural information processing systems},
  volume={30},
  year={2017}
}

@article{bib25,
  title={Long Short-term Memory},
  author={Hochreiter, S},
  journal={Neural Computation MIT-Press},
  address = {Cambridge, MA},
  year={1997}
}

@article{bib26,
  title={Long short-term memory recurrent neural network architectures for large scale acoustic modeling},
  author={Sak, Hasim and Senior, Andrew W and Beaufays, Fran{\c{c}}oise},
  year={2014}
}

@misc{bib27,
      title={Learning Phrase Representations using RNN Encoder-Decoder for Statistical Machine Translation}, 
      author={Kyunghyun Cho and Bart van Merrienboer and Caglar Gulcehre and Dzmitry Bahdanau and Fethi Bougares and Holger Schwenk and Yoshua Bengio},
      year={2014},
      eprint={1406.1078},
      archivePrefix={arXiv},
      primaryClass={cs.CL},
      url={https://arxiv.org/abs/1406.1078}, 
}

@misc{bib28,
      title={Empirical Evaluation of Gated Recurrent Neural Networks on Sequence Modeling}, 
      author={Junyoung Chung and Caglar Gulcehre and KyungHyun Cho and Yoshua Bengio},
      year={2014},
      eprint={1412.3555},
      archivePrefix={arXiv},
      primaryClass={cs.NE},
      url={https://arxiv.org/abs/1412.3555}, 
}

@article{bib29,
  title={Support-Vector Networks},
  author={Cortes, Corinna},
  journal={Machine Learning},
  year={1995}
}

@article{bib30,
  title={Statistical learning theory},
  author={Vapnik, Vladimir},
  journal={John Wiley \& Sons google schola},
  volume={2},
  pages={831--842},
  year={1998}
}

@article{bib31,
  title={Credit scoring with a data mining approach based on support vector machines},
  author={Huang, Cheng-Lung and Chen, Mu-Chen and Wang, Chieh-Jen},
  journal={Expert systems with applications},
  volume={33},
  number={4},
  pages={847--856},
  year={2007},
  publisher={Elsevier}
}

@book{bib32,
  title={Applied logistic regression analysis},
  author={Menard, Scott},
  volume={106},
  year={2002},
  address = {Thousand Oaks, CA},
  publisher={Sage}
}

@book{bib33,
  title={The elements of statistical learning: data mining, inference, and prediction},
  author={Hastie, Trevor and Tibshirani, Robert and Friedman, Jerome},
  year={2009},
  address = {New York},
  publisher={Springer Science \& Business Media}
}

@book{bib34,
  title={Pattern recognition and machine learning},
  author={Bishop, Christopher M and Nasrabadi, Nasser M},
  volume={4},
  number={4},
  year={2006},
  address = {New York},
  publisher={Springer}
}

@article{bib35,
  title={Bayesian inference for categorical data analysis},
  author={Agresti, Alan and Hitchcock, David B},
  journal={Statistical Methods and Applications},
  volume={14},
  pages={297--330},
  year={2005},
  address = {New York},
  publisher={Springer}
}

@book{bib36,
  title={Machine learning: a probabilistic perspective},
  author={Murphy, Kevin P},
  year={2012},
  address = {Cambridge, MA},
  publisher={MIT press}
}

@article{bib37,
  title={Random forests},
  author={Breiman, Leo},
  journal={Machine learning},
  volume={45},
  pages={5--32},
  year={2001},
  address = {New York},
  publisher={Springer}
}

@article{bib38,
  title={Classification and regression by randomForest},
  author={Liaw, A},
  journal={R news},
  year={2002}
}

@article{bib39,
  title={A Baseline Drift-Elimination Algorithm for Strain Measurement-System Signals Based on the Transformer Model},
  author={Wang, Yusen and Zhang, Lei and Qi, Xue and Yang, Xiaopeng and Tan, Qiulin},
  journal={Applied Sciences},
  volume={14},
  number={11},
  pages={4447},
  year={2024},
  publisher={MDPI}
}

@inproceedings{bib40,
  title={Comparative Analysis of Transformer based Models for Question Answering},
  author={Rawat, Anchal and Samant, Surender Singh},
  booktitle={2022 2nd International Conference on Innovative Sustainable Computational Technologies (CISCT)},
  pages={1--6},
  year={2022},
  organization={IEEE}
}

@article{bib41,
  title={Expanding the prediction capacity in long sequence time-series forecasting},
  author={Zhou, Haoyi and Li, Jianxin and Zhang, Shanghang and Zhang, Shuai and Yan, Mengyi and Xiong, Hui},
  journal={Artificial Intelligence},
  volume={318},
  pages={103886},
  year={2023},
  publisher={Elsevier}
}

@article{bib42,
  title={Transformers in time series: A survey},
  author={Wen, Qingsong and Zhou, Tian and Zhang, Chaoli and Chen, Weiqi and Ma, Ziqing and Yan, Junchi and Sun, Liang},
  journal={arXiv preprint arXiv:2202.07125},
  year={2022}
}

@inproceedings{bib43,
  title={Comparative Analysis of Transformer based Models for Question Answering},
  author={Rawat, Anchal and Samant, Surender Singh},
  booktitle={2022 2nd International Conference on Innovative Sustainable Computational Technologies (CISCT)},
  pages={1--6},
  year={2022},
  organization={IEEE}
}

@article{bib44,
  title={Transformers for scientific data: a pedagogical review for astronomers},
  author={Tanoglidis, Dimitrios and Jain, Bhuvnesh and Qu, Helen},
  journal={arXiv preprint arXiv:2310.12069},
  year={2023}
}

@article{bib45,
  title={Logistic regression analysis of populations of electrophysiological models to assess proarrythmic risk},
  author={Morotti, Stefano and Grandi, Eleonora},
  journal={MethodsX},
  volume={4},
  pages={25--34},
  year={2017},
  publisher={Elsevier}
}

@article{bib46,
  title={Support vector machines to detect physiological patterns for EEG and EMG-based human--computer interaction: a review},
  author={Quitadamo, LR and Cavrini, F and Sbernini, L and Riillo, F and Bianchi, L and Seri, S and Saggio, G},
  journal={Journal of neural engineering},
  volume={14},
  number={1},
  pages={011001},
  year={2017},
  publisher={IOP Publishing}
}

@article{bib47,
  title={Random forest-based approach for physiological functional variable selection for driver’s stress level classification},
  author={El Haouij, Neska and Poggi, Jean-Michel and Ghozi, Raja and Sevestre-Ghalila, Sylvie and Ja{\"\i}dane, M{\'e}riem},
  journal={Statistical Methods \& Applications},
  volume={28},
  pages={157--185},
  year={2019},
  address = {New York},
  publisher={Springer}
}

@article{bib48,
  title={Multidisciplinary pattern recognition applications: A review},
  author={Paolanti, Marina and Frontoni, Emanuele},
  journal={Computer Science Review},
  volume={37},
  pages={100276},
  year={2020},
  publisher={Elsevier}
}

@article{bib49,
  title={A structured approach to predictive modeling of a two-class problem using multidimensional data sets},
  author={Spratt, Heidi and Ju, Hyunsu and Brasier, Allan R},
  journal={Methods},
  volume={61},
  number={1},
  pages={73--85},
  year={2013},
  publisher={Elsevier}
}

@article{bib50,
  title={Working-class jobs and new parents' mental health},
  author={Perry-Jenkins, Maureen and Smith, JuliAnna Z and Goldberg, Abbie E and Logan, Jade},
  journal={Journal of Marriage and Family},
  volume={73},
  number={5},
  pages={1117--1132},
  year={2011},
  publisher={Wiley Online Library}
}

@article{bib51,
  title={Workplace policies and mental health among working-class, new parents},
  author={Perry-Jenkins, Maureen and Smith, JuliAnna Z and Wadsworth, Lauren Page and Halpern, Hillary Paul},
  journal={Community, work \& family},
  volume={20},
  number={2},
  pages={226--249},
  year={2017},
  publisher={Taylor \& Francis}
}

@article{bib52,
  title={Advanced Machine Learning for Financial Markets: A PCA-GRU-LSTM Approach},
  author={Liu, Bingchun and Lai, Mingzhao},
  journal={Journal of the Knowledge Economy},
  pages={1--35},
  year={2024},
  address = {New York},
  publisher={Springer}
}

@article{fu2023decoder,
  title={Decoder-only or encoder-decoder? interpreting language model as a regularized encoder-decoder},
  author={Fu, Zihao and Lam, Wai and Yu, Qian and So, Anthony Man-Cho and Hu, Shengding and Liu, Zhiyuan and Collier, Nigel},
  journal={arXiv preprint arXiv:2304.04052},
  year={2023}
}

@article{gao2022encoder,
  title={Is Encoder-Decoder Redundant for Neural Machine Translation?},
  author={Gao, Yingbo and Herold, Christian and Yang, Zijian and Ney, Hermann},
  journal={arXiv preprint arXiv:2210.11807},
  year={2022}
}

@article{atkinson1971control,
  title={The control of short-term memory},
  author={Atkinson, Richard C and Shiffrin, Richard M},
  journal={Scientific american},
  volume={225},
  number={2},
  pages={82--91},
  year={1971},
  publisher={JSTOR}
}

@inproceedings{agarwal2023physiological,
  title={Physiological Signals based Student Grades Prediction using Machine Learning},
  author={Agarwal, Vanshikha and Ahmad, Naeem and Hasan, Md Gulzarul},
  booktitle={2023 OITS International Conference on Information Technology (OCIT)},
  pages={208--213},
  year={2023},
  organization={IEEE}
}

@article{lazarou2024predicting,
  title={Predicting stress levels using physiological data: Real-time stress prediction models utilizing wearable devices},
  author={Lazarou, Evgenia and Exarchos, Themis P},
  journal={AIMS Neuroscience},
  volume={11},
  number={2},
  pages={76--102},
  year={2024}
}

@inproceedings{wijsman2011towards,
  title={Towards mental stress detection using wearable physiological sensors},
  author={Wijsman, Jacqueline and Grundlehner, Bernard and Liu, Hao and Hermens, Hermie and Penders, Julien},
  booktitle={2011 Annual International Conference of the IEEE Engineering in Medicine and Biology Society},
  pages={1798--1801},
  year={2011},
  organization={IEEE}
}

@article{van2020review,
  title={A review on the long short-term memory model},
  author={Van Houdt, Greg and Mosquera, Carlos and N{\'a}poles, Gonzalo},
  journal={Artificial Intelligence Review},
  volume={53},
  number={8},
  pages={5929--5955},
  year={2020},
  address = {New York},
  publisher={Springer}
}

@article{mienye2024recurrent,
  title={Recurrent neural networks: A comprehensive review of architectures, variants, and applications},
  author={Mienye, Ibomoiye Domor and Swart, Theo G and Obaido, George},
  journal={Information},
  volume={15},
  number={9},
  pages={517},
  year={2024},
  publisher={MDPI}
  }

@article{shiri2023comprehensive,
  title={A comprehensive overview and comparative analysis on deep learning models: CNN, RNN, LSTM, GRU},
   author={Shiri, Farhad Mortezapour and Perumal, Thinagaran and Mustapha, Norwati and Mohamed, Raihani},
   journal={arXiv preprint arXiv:2305.17473},
   year={2023}
 }

@ article{nazareth2023financial,
  title={Financial applications of machine learning: A literature review},
  author={Nazareth, Noella and Reddy, Yeruva Venkata Ramana},
  journal={Expert Systems with Applications},
  volume={219},
  pages={119640},
  year={2023},
  publisher={Elsevier}
}

@article{rithani2023review,
  title={A review on big data based on deep neural network approaches},
  author={Rithani, M and Kumar, R Prasanna and Doss, Srinath},
  journal={Artificial Intelligence Review},
  volume={56},
  number={12},
  pages={14765--14801},
  year={2023},
  address = {New York},
  publisher={Springer}}

@article{ahmad2023framework,
  title={A framework to estimate cognitive load using physiological data},
  author={Ahmad, Muneeb Imtiaz and Keller, Ingo and Robb, David A and Lohan, Katrin S},
  journal={Personal and Ubiquitous Computing},
  pages={1--15},
  year={2023},
  address = {New York},
  publisher={Springer}
}

@article{martinez2021review,
  title={A review of machine learning in hypertension detection and blood pressure estimation based on clinical and physiological data},
  author={Martinez-R{\'\i}os, Erick and Montesinos, Luis and Alfaro-Ponce, Mariel and Pecchia, Leandro},
  journal={Biomedical Signal Processing and Control},
  volume={68},
  pages={102813},
  year={2021},
  publisher={Elsevier}
}

@article{bhandari2022predicting,
  title={Predicting stock market index using LSTM},
  author={Bhandari, Hum Nath and Rimal, Binod and Pokhrel, Nawa Raj and Rimal, Ramchandra and Dahal, Keshab R and Khatri, Rajendra KC},
  journal={Machine Learning with Applications},
  volume={9},
  pages={100320},
  year={2022},
  publisher={Elsevier}
}

@article{rimal2023comparative,
  title={Comparative study of various machine learning methods on ASD classification},
  author={Rimal, Ramchandra and Brannon, Mitchell and Wang, Yingxin and Yang, Xin},
  journal={International Journal of Data Science and Analytics},
  pages={1--15},
  year={2023},
  address = {New York},
  publisher={Springer}
}

@article{bhandari2024implementation,
  title={Implementation of deep learning models in predicting ESG index volatility},
  author={Bhandari, Hum Nath and Pokhrel, Nawa Raj and Rimal, Ramchandra and Dahal, Keshab R and Rimal, Binod},
  journal={Financial Innovation},
  volume={10},
  number={1},
  pages={75},
  year={2024},
  address = {New York},
  publisher={Springer}
}

@article{rimal2024real,
  title={Real Estate Market Prediction Using Deep Learning Models},
  author={Rimal, Ramchandra and Rimal, Binod and Bhandari, Hum Nath and Pokhrel, Nawa Raj and Dahal, Keshab R},
  journal={Annals of Data Science},
  pages={1--44},
  year={2024},
  address = {New York},
  publisher={Springer}
}

@article{rimal2024identifying,
  title={Identifying the neurocognitive difference between two groups using supervised learning},
  author={Rimal, Ramchandra},
  journal={Statistics, Optimization \& Information Computing},
  volume={12},
  number={1},
  pages={15--33},
  year={2024}
}

\newpage

\begin{appendices}

\section{Statistical Analysis Results}\label{secA1}
This section contains the results on statistical analysis of the performance across 30 replicates on the test data.

\subsection{ Hypothesis Testing Results}\label{subsecA2}
The results of the Welch's t-test  for each evaluation metric are reported in Tables~\ref{tab:ttest_accuracy}
- ~\ref{tab:ttest_f1}.
%%%%%%  Paramateric Tests %%%%%%%%%%%%%%%%%%
\begin{table}[H]
\caption{Welch's t-test Results for Accuracy}
\label{tab:ttest_accuracy}
\begin{tabular}{lllrrl}
\toprule
Metric & Model 1 & Model 2 & t-statistic & p-value & Conclusion \\
\midrule
Accuracy & GRU & LSTM & -3.3509 & 0.0016 & Significant \\
Accuracy & GRU & Logistic Regression & 168.4896 & 0.0000 & Significant \\
Accuracy & GRU & Random Forest & -13.9247 & 0.0000 & Significant \\
Accuracy & GRU & Transformer & 4.4621 & 0.0001 & Significant \\
Accuracy & LSTM & Logistic Regression & 276.7226 & 0.0000 & Significant \\
Accuracy & LSTM & Random Forest & -16.2388 & 0.0000 & Significant \\
Accuracy & LSTM & Transformer & 11.2070 & 0.0000 & Significant \\
Accuracy & Logistic Regression & Random Forest & -1435.8238 & 0.0000 & Significant \\
Accuracy & Logistic Regression & Transformer & -315.0201 & 0.0000 & Significant \\
Accuracy & Random Forest & Transformer & 37.2586 & 0.0000 & Significant \\
\bottomrule
\end{tabular}
\end{table}
%%%%%%%%%%%%%%%%%%%
\begin{table}[H]
\caption{Welch's t-test Results for Precision}
\label{tab:ttest_precision}
\begin{tabular}{lllrrl}
\toprule
Metric & Model 1 & Model 2 & t-statistic & p-value & Conclusion \\
\midrule
Precision & GRU & LSTM & -3.3430 & 0.0016 & Significant \\
Precision & GRU & Logistic Regression & 185.1526 & 0.0000 & Significant \\
Precision & GRU & Random Forest & -14.2745 & 0.0000 & Significant \\
Precision & GRU & Transformer & 4.6644 & 0.0000 & Significant \\
Precision & LSTM & Logistic Regression & 296.6288 & 0.0000 & Significant \\
Precision & LSTM & Random Forest & -16.4847 & 0.0000 & Significant \\
Precision & LSTM & Transformer & 11.2800 & 0.0000 & Significant \\
Precision & Logistic Regression & Random Forest & -1312.6440 & 0.0000 & Significant \\
Precision & Logistic Regression & Transformer & -333.7955 & 0.0000 & Significant \\
Precision & Random Forest & Transformer & 37.2911 & 0.0000 & Significant \\
\bottomrule
\end{tabular}
\end{table}
%%%%%%%%%%%%%%%%%%
\begin{table}[H]
\caption{Welch's t-test Results for Recall}
\label{tab:ttest_recall}
\begin{tabular}{lllrrl}
\toprule
Metric & Model 1 & Model 2 & t-statistic & p-value & Conclusion \\
\midrule
Recall & GRU & LSTM & -3.3509 & 0.0016 & Significant \\
Recall & GRU & Logistic Regression & 168.4896 & 0.0000 & Significant \\
Recall & GRU & Random Forest & -13.9247 & 0.0000 & Significant \\
Recall & GRU & Transformer & 4.4563 & 0.0001 & Significant \\
Recall & LSTM & Logistic Regression & 276.7226 & 0.0000 & Significant \\
Recall & LSTM & Random Forest & -16.2388 & 0.0000 & Significant \\
Recall & LSTM & Transformer & 11.1996 & 0.0000 & Significant \\
Recall & Logistic Regression & Random Forest & -1435.8238 & 0.0000 & Significant \\
Recall & Logistic Regression & Transformer & -315.0975 & 0.0000 & Significant \\
Recall & Random Forest & Transformer & 37.2535 & 0.0000 & Significant \\
\bottomrule
\end{tabular}
\end{table}
%%%%%%%%%%%%%%%%%%%%%
\begin{table}[H]
\caption{Welch's t-test Results for F1}
\label{tab:ttest_f1}
\begin{tabular}{lllrrl}
\toprule
Metric & Model 1 & Model 2 & t-statistic & p-value & Conclusion \\
\midrule
F1 & GRU & LSTM & -3.3489 & 0.0016 & Significant \\
F1 & GRU & Logistic Regression & 177.9488 & 0.0000 & Significant \\
F1 & GRU & Random Forest & -13.9588 & 0.0000 & Significant \\
F1 & GRU & Transformer & 4.4622 & 0.0001 & Significant \\
F1 & LSTM & Logistic Regression & 291.0075 & 0.0000 & Significant \\
F1 & LSTM & Random Forest & -16.3609 & 0.0000 & Significant \\
F1 & LSTM & Transformer & 11.2279 & 0.0000 & Significant \\
F1 & Logistic Regression & Random Forest & -1238.4162 & 0.0000 & Significant \\
F1 & Logistic Regression & Transformer & -330.2115 & 0.0000 & Significant \\
F1 & Random Forest & Transformer & 37.3808 & 0.0000 & Significant \\
\bottomrule
\end{tabular}
\end{table}

The results associated with  the Mann–Whitney U test for each evaluation metric are reported in Tables~\ref{tab:mannwhitney_accuracy}  - ~\ref{tab:mannwhitney_F1}.
%%%%%% Non - Paramateric Tests %%%%%%%%%%%%%%%%%%
\begin{table}[H]
\caption{Mann-Whitney U test Results for Accuracy}
\label{tab:mannwhitney_accuracy}
\begin{tabular}{lllrrl}
\toprule
Metric & Model 1 & Model 2 & U-statistic & p-value & Conclusion \\
\midrule
Accuracy & GRU & LSTM & 219.0000 & 0.0007 & Significant \\
Accuracy & GRU & Logistic Regression & 900.0000 & 0.0000 & Significant \\
Accuracy & GRU & Random Forest & 0.0000 & 0.0000 & Significant \\
Accuracy & GRU & Transformer & 754.0000 & 0.0000 & Significant \\
Accuracy & LSTM & Logistic Regression & 900.0000 & 0.0000 & Significant \\
Accuracy & LSTM & Random Forest & 0.0000 & 0.0000 & Significant \\
Accuracy & LSTM & Transformer & 871.0000 & 0.0000 & Significant \\
Accuracy & Logistic Regression & Random Forest & 0.0000 & 0.0000 & Significant \\
Accuracy & Logistic Regression & Transformer & 0.0000 & 0.0000 & Significant \\
Accuracy & Random Forest & Transformer & 900.0000 & 0.0000 & Significant \\
\bottomrule
\end{tabular}
\end{table}
%%%%%%%%%%%%%%%%%%%

\begin{table}[H]
\caption{Mann-Whitney U test Results for Precision}
\label{tab:mannwhitney_precision}
\begin{tabular}{lllrrl}
\toprule
Metric & Model 1 & Model 2 & U-statistic & p-value & Conclusion \\
\midrule
Precision & GRU & LSTM & 225.0000 & 0.0009 & Significant \\
Precision & GRU & Logistic Regression & 900.0000 & 0.0000 & Significant \\
Precision & GRU & Random Forest & 0.0000 & 0.0000 & Significant \\
Precision & GRU & Transformer & 753.0000 & 0.0000 & Significant \\
Precision & LSTM & Logistic Regression & 900.0000 & 0.0000 & Significant \\
Precision & LSTM & Random Forest & 0.0000 & 0.0000 & Significant \\
Precision & LSTM & Transformer & 871.0000 & 0.0000 & Significant \\
Precision & Logistic Regression & Random Forest & 0.0000 & 0.0000 & Significant \\
Precision & Logistic Regression & Transformer & 0.0000 & 0.0000 & Significant \\
Precision & Random Forest & Transformer & 900.0000 & 0.0000 & Significant \\
\bottomrule
\end{tabular}
\end{table}
%%%%%%%%%%%%%%%%%%%%%%%%%%%
\begin{table}[H]
\caption{Mann-Whitney U test Results for Recall}
\label{tab:mannwhitney_recall}
\begin{tabular}{lllrrl}
\toprule
Metric & Model 1 & Model 2 & U-statistic & p-value & Conclusion \\
\midrule
Recall & GRU & LSTM & 219.0000 & 0.0007 & Significant \\
Recall & GRU & Logistic Regression & 900.0000 & 0.0000 & Significant \\
Recall & GRU & Random Forest & 0.0000 & 0.0000 & Significant \\
Recall & GRU & Transformer & 754.0000 & 0.0000 & Significant \\
Recall & LSTM & Logistic Regression & 900.0000 & 0.0000 & Significant \\
Recall & LSTM & Random Forest & 0.0000 & 0.0000 & Significant \\
Recall & LSTM & Transformer & 871.0000 & 0.0000 & Significant \\
Recall & Logistic Regression & Random Forest & 0.0000 & 0.0000 & Significant \\
Recall & Logistic Regression & Transformer & 0.0000 & 0.0000 & Significant \\
Recall & Random Forest & Transformer & 900.0000 & 0.0000 & Significant \\
\bottomrule
\end{tabular}
\end{table}
%%%%%%%%%%%%%%%%%%%%%%%%%%%%%

\begin{table}[H] 
\caption{Mann-Whitney U test Results for F1}
\label{tab:mannwhitney_F1}
\begin{tabular}{lllrrl}
\toprule
Metric & Model 1 & Model 2 & U-statistic & p-value & Conclusion \\
\midrule
F1 & GRU & LSTM & 222.0000 & 0.0008 & Significant \\
F1 & GRU & Logistic Regression & 900.0000 & 0.0000 & Significant \\
F1 & GRU & Random Forest & 0.0000 & 0.0000 & Significant \\
F1 & GRU & Transformer & 754.0000 & 0.0000 & Significant \\
F1 & LSTM & Logistic Regression & 900.0000 & 0.0000 & Significant \\
F1 & LSTM & Random Forest & 0.0000 & 0.0000 & Significant \\
F1 & LSTM & Transformer & 871.0000 & 0.0000 & Significant \\
F1 & Logistic Regression & Random Forest & 0.0000 & 0.0000 & Significant \\
F1 & Logistic Regression & Transformer & 0.0000 & 0.0000 & Significant \\
F1 & Random Forest & Transformer & 900.0000 & 0.0000 & Significant \\
\bottomrule
\end{tabular}
\end{table}

\end{appendices}

\end{document}